\documentclass{article}

\usepackage[preprint]{neurips_2024}

\usepackage[utf8]{inputenc} 
\usepackage[T1]{fontenc}    
\usepackage{hyperref}       
\usepackage{url}            
\usepackage{booktabs}       
\usepackage{amsfonts}       
\usepackage{nicefrac}       
\usepackage{microtype}      
\usepackage{xcolor}         

\usepackage{multirow}
\usepackage{graphicx}
\usepackage{wrapfig}  
\usepackage{amsmath}
\usepackage{bm}
\usepackage{amssymb}

\usepackage{graphicx}
\usepackage{subcaption}
\usepackage{ulem}
\usepackage{threeparttable}
\usepackage{enumitem}
\usepackage{ulem}

\usepackage{algorithm}
\usepackage{algorithmic}
\usepackage{tcolorbox}
\usepackage{listings}
\usepackage{xcolor}

\definecolor{brightsalmon}{RGB}{255, 110, 96}
\definecolor{RoyalBlue}{rgb}{0.25, 0.41, 0.88}
\hypersetup{
    colorlinks = true,
    urlcolor = RoyalBlue,  
}

\title{Large Language Models are Interpretable Learners}

\author{%
  Ruochen Wang\thanks{Work completed during internship at Google.} \\
  UCLA \\
  \And
  Si Si \\
  Google Research \\
  \And
  Felix Yu \\
  Google Research \\
  \And
  Dorothea Wiesmann \\
  Google Research \\
  \And
  Cho-Jui Hsieh \\
  Google Research \\
  \And
  Inderjit Dhillon \\
  Google Research
}

\begin{document}

\maketitle

\begingroup
\centering
\vspace{-8mm}

\url{https://github.com/ruocwang/llm-symbolic-program}

\vspace{+8mm}
\par
\endgroup

\begin{abstract}
The trade-off between expressiveness and interpretability remains a core challenge when building human-centric predictive models for classification and decision-making. While symbolic rules offer interpretability, they often lack expressiveness, whereas neural networks excel in performance but are known for being black boxes. In this paper, we show a combination of Large Language Models (LLMs) and symbolic programs can bridge this gap. In the proposed LLM-based Symbolic Programs (LSPs), the pretrained LLM with natural language prompts provides a massive set of interpretable modules that can transform raw input into natural language concepts. Symbolic programs then integrate these modules into an interpretable decision rule. To train LSPs, we develop a divide-and-conquer approach to incrementally build the program from scratch, where the learning process of each step is guided by LLMs. To evaluate the effectiveness of LSPs in extracting interpretable and accurate knowledge from data, we introduce IL-Bench, a collection of diverse tasks, including both synthetic and real-world scenarios across different modalities. Empirical results demonstrate LSP's superior performance compared to traditional neurosymbolic programs and vanilla automatic prompt tuning methods. Moreover, as the knowledge learned by LSP is a combination of natural language descriptions and symbolic rules, it is easily transferable to humans (interpretable), and other LLMs, and generalizes well to out-of-distribution samples.
\end{abstract}

\vspace{-7pt}
\section{Introduction}
\label{sec.intro}
\vspace{-5pt}
Learning interpretable predictive models from annotated data remains a key challenge in human-centric AI.
Given input-output pairs $\{(x_i, y_i)\}$, the objective is to learn a function $f: x\rightarrow y$ that not only fits the data accurately but is also interpretable.
 tIn this context, a strong form of "interpretable" means that individuals with no prior domain knowledge can understand and apply the decision rules demonstrated by $f$, facilitating \textit{the transfer of knowledge from AI to humans}.
This is crucial not only for enhancing the transparency of AI systems but also for enabling humans to learn from these models, empowering various human-in-the-loop applications such as scientific discovery, material synthesis, and automatic data annotation~\citep{nsp}.

\begin{figure}[t]
\vspace{-15pt}
    \centering
    \includegraphics[width=0.95\textwidth, height=2.2in]{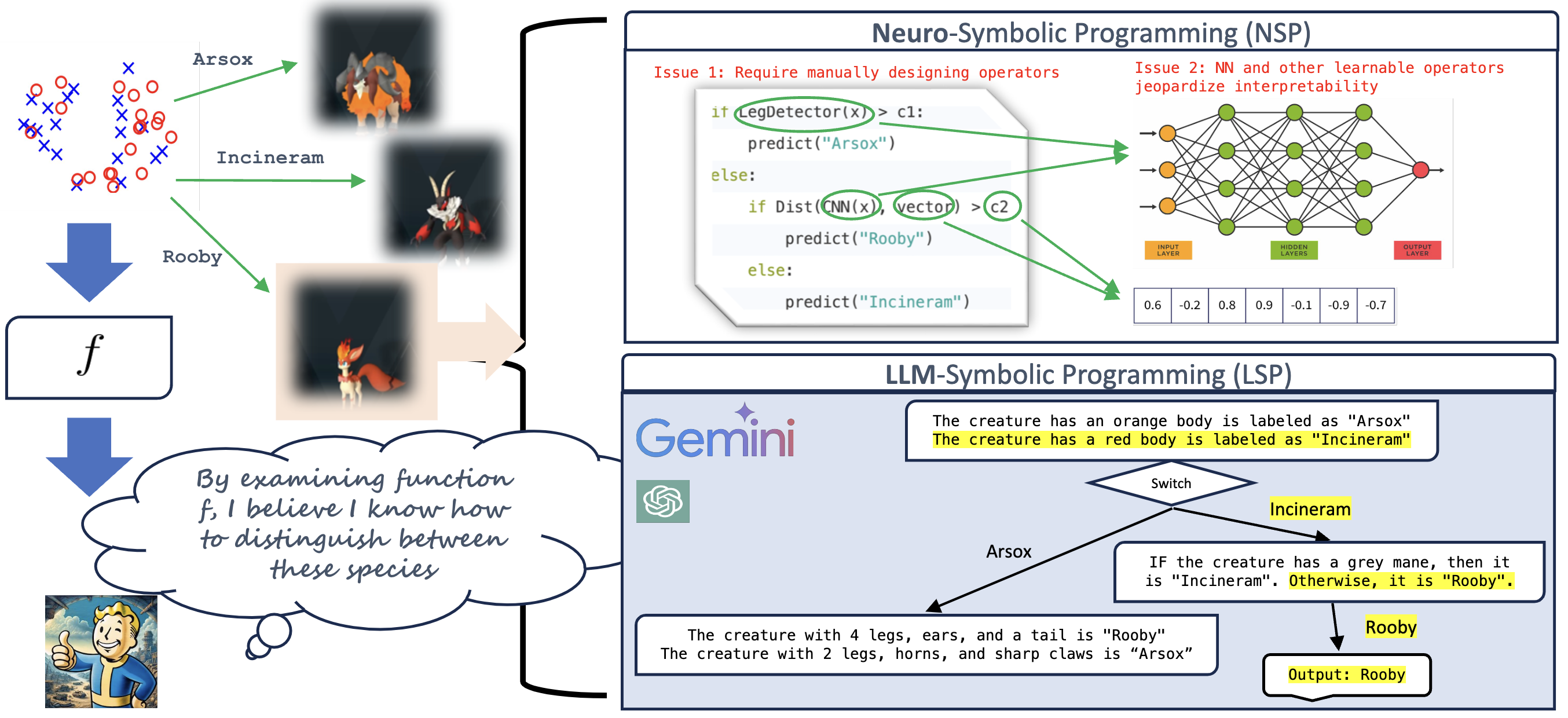}
    \caption{
        \small
        \textbf{Illustration of LLM-Symbolic vs. Neuro-Symbolic Program on interpretable learning task.}
        The goal is to develop a model that allows humans with no prior knowledge to replicate AI’s decisions by following the same rules as the model. While \textbf{NSP (Top right)} offers a certain level of interpretability, it heavily relies on manually designing operators, and the inclusion of neural operators often reduces interpretability. In contrast, \textbf{LSP (Bottom right)} generates fully interpretable programs with the help of versatile LLM modules.
        \vspace{-15pt}
    }
    \label{fig:inference}
\end{figure}

Consider an exemplar task of classifying species in Palworld~\citep{palworld} - a newly released Pokemon-style game - based on a few image-label pairs, as illustrated in Figure~\ref{fig:inference}.
The ultimate goal is that even humans unfamiliar with Palworld can replicate AI's decisions by following the same predictive rules after examining the model trained on the data.
This task effectively represents the challenge of extracting interpretable knowledge, such as species characteristics, from data.
The algorithm we propose in this paper learns a model following the decision rule illustrated in Figure~\ref{fig:inference}, which is designed to be easily understood and reproduced by humans. In essence, this problem can be viewed as discovering interpretable knowledge (e.g., the properties of a species in Palworld) from the data.

Despite extensive research, the problem of developing a fully interpretable predictive model has not been fully addressed.
Traditional methods often face a trade-off between expressiveness and interpretability:
Deep neural networks, for instance, are powerful yet operate as "black boxes".
Although post-hoc explanation methods attempt to make these models more transparent by identifying influential features~\citep{lime}, they do not clarify the underlying decision-making processes and have no control over the learning process.
Directly learning interpretable models like (locally) linear, tree-based often falls short in expressiveness, especially with complex inputs like images.

To address this challenge, \textbf{Neurosymbolic Programs (NSPs)}~\citep{nsp, near, dpad, prototree} offer a promising solution by modeling the decision rule as a program incorporating both symbolic operations and neural network modules.
Despite this, the inherent trade-off between expressiveness and interpretability persists.
While the integration of neural modules enhances expressiveness, it also compromises the program's overall interpretability.
Additionally, designing effective symbolic operators requires significant expertise and is critical for the performance of the resulting program, necessitating careful customization for each specific dataset~\citep{nsp, near, dpad}.

Is it possible to harness the power of neural networks within Neurosymbolic Programs without compromising interpretability?
This paper presents an affirmative answer.
Our key insight is that (Multimodal) LLMs encompass a variety of powerful, conditional probabilistic sub-models.
These models share a unified parametric architecture with the unconditional parent LLM (Super Model), yet distinctive defined by their respective prompts.
Therefore, crafting prompts (by either Human or meta-LLMs) for LLM is equivalent to searching over the hypothesis space spanned by these submodels.
This yields an infinite set of neural network-based operations that are inherently interpretable and can serve as fundamental ``learnable'' building blocks within Neurosymbolic Programs.

Building on this insight, we introduce a novel framework termed \textbf{LLM-Symbolic Programs (LSPs)}, defined and learned through LLMs.
Our approach leverages a minimal Domain-Specific Language (DSL) set with only two operators: prompted-LLM and conditional branching, yielding a classic decision-making process structured as trees. We then propose a learning algorithm to incrementally learn the tree using LLMs with prompt optimization. 
To thoroughly evaluate the efficacy of LSPs, we construct the \textbf{Interpretable-Learning-Benchmark} of diverse predictive tasks, containing both synthetic and real-world data across vision and text modalities.
Our empirical findings show that LSPs surpass the accuracy of both traditional XAI methods and LLMs prompted with automatically learned instructions, all while maintaining human interpretability.
These results highlight the potential of LSPs to significantly enhance the performance and utility of Multimodal LLMs in various applications.


\vspace{-5pt}
\section{LLM-Symbolic Programs}
\label{sec.method.algo}
\vspace{-10pt}
This section explains our proposed framework:  LLM-Symbolic Programs.
Section~\ref{sec:method.prelim} reviews Neurosymbolic Learning method.
Section~\ref{sec:method.lsp} discusses utilizing LLM to implement interpretable programs, including a connection between prompted-LLM and interpretable unit (Section~\ref{sec:method.lsp.theory}), the Domain Specific Language (Section~\ref{sec:method.lsp.dsl}) and learning algorithm (Section~\ref{sec:method.lsp.algo}).

\vspace{-5pt}
\subsection{Preliminaries on Neurosymbolic Learning}
\label{sec:method.prelim}
\vspace{-5pt}
    NeuroSymbolic Programmings (NSPs)~\citep{nsp, near, dpad, distill} seek to couple classical symbolic methods with contemporary neural networks to build expressive and interpretable models.
    The learning method for NSPs is often characterized by (1) a \textbf{Domain Specific Language (DSL)} that specifies available operations of the program and (2) \textbf{a learning algorithm} for finding the best instance.
    The resulting programs are structured, neuro-symbolic terms that follow the syntax specified by the DSL.
    
    \paragraph{Domain-Specific Language (DSL)}
    DSL in NSPs comprises manually defined operators, including interpretable symbolic (e.g. \texttt{if-then-else}) and expressive neural components (e.g. \texttt{cnn(x, $\theta$)}).
    These operators can be chained to construct various tree-structured programs, a.k.a. computation graphs.
    Eq~\eqref{eq:nsp_dsl} presents an example DSL used to construct the  program for predicting the creature species in Figure~\ref{fig:inference}.
    Here, $x$ and $c$ represents inputs and constants, and $\alpha$ denotes a sub-program:
    \begin{align}
        \alpha ::= x \mid c
        \mid \texttt{Add}(\alpha_1, \alpha_2) \mid \texttt{Mul}(\alpha_1, \alpha_2)
        \mid \text{If } \alpha_1 \text{ Then } \alpha_2 \text{ Else } \alpha_3
        \mid \texttt{cnn}(x, \theta) \mid \texttt{Dist}(\alpha_1, \alpha_2).
    \label{eq:nsp_dsl}
    \end{align}
    
    \paragraph{Co-optimization of program structure and learnable parameters}
    In NSPs, the construction of a program involves solving a combinatorial optimization problem for both the program structure and the parameters of its learnable operators (e.g. neural components). As the number of DSL operators increases, the complexity of this task grows exponentially. To make the search process more tractable, existing research employs various approximation techniques to efficiently identify viable candidates, including greedy tree search~\citep{near}, continuous relaxation~\citep{dpad}, distillation~\citep{distill} and meta-learning~\citep{nsp}.
    
    \paragraph{Limitations}
While the integration of symbolic and neural components in NSPs represents a promising innovation, the incorporating of neural modules inevitably introduces black-box components and makes the program  non-interpretable. Researchers have attempted to address this issue through two primary approaches: restricting the DSL to only interpretable operators~\citep{near, dpad}, or employing prototype learning to derive relatively interpretable neural modules~\citep{prototree, prototype1, prototype2}. However, the DSL approach is not automatic, heavily relies on domain expertise, and potentially overlooking crucial information not identified by experts; Conversely, prototype learning aims to represent the concept of each neural module by a set of representative samples, which is not guaranteed to success.

\subsection{LLM-Symbolic Programs}
\label{sec:method.lsp}
This section explores how LLMs can effectively be utilized to implement NSPs' modules that are expressive, interpretable, and straightforward to learn with LLMs.

    \subsubsection{Prompted-LLM as an interpretable unit}
    \label{sec:method.lsp.theory}
    The trade-off between interpretability and expressiveness presents a fundamental limitation in machine learning. Machines perceive images and text as raw binary signals, and transforming these into interpretable concepts inevitably requires complex and non-interpretable components, such as neural networks. Even human perception remains non-interpretable, as we lack a complete understanding of how the brain processes signals.
    However, The following analysis suggests that pretrained LLM offer a potential avenue to bridge this gap.
    
    \paragraph{Connection between interpretable learning and prompting}
    LLMs pretrained on the next-token prediction task model the following joint distribution of a sequence of tokens $\{w_t\}_{t=1}^T$
    \begin{align*}
        P(w_1, w_2, \ldots, w_T) = \prod\nolimits_{t=1}^T P(w_t \mid w_{t-1}, w_{t-2}, \dots, 1) = f_\theta(w_t \mid w_1, w_2, \ldots, w_{t - 1}),
    \end{align*}
    where the conditional probabilities are parameterized by an auto-regressive model $f(\cdot;\theta)$ (e.g. Transformer) and each word $w_t$ is predicted given all the preceding tokens.
    The pretraining objective minimizes the following negative log-likelihood:
    \begin{align}
        \min_\theta \mathcal{L}(\theta) = -\sum\nolimits_{t=1}^T \log f_\theta(w_t \mid w_{t-1}, \dots, w_1).
        \label{eq:train}
    \end{align}

    A key observation from Eq.~\eqref{eq:train} is that the training process optimizes a ``SuperNet" of conditional probabilistic models (CPM), each defined by an instruction $s$: $f_{s, \theta}(y|x) = f_\theta(y \mid x, s)$, where $x$ is the input and $s$ is the instruction for a particular task. Therefore, with a fixed LLM, the set of natural language prompts, denoted as $\cal{S}$, provides a massive set of interpretable neural network modules for the task.
    For a given dataset $\{(x_i, y_i)\}_{i=1}^n$, finding the best prompt  to minimize the empirical loss, 
        $\min_{s\in {\cal S}} \sum_{i=1}^n{\cal L}((f_{s, \theta}(y_i \mid x_i)))$, 
    can be viewed as a form of learning, and the resulting model is inherently interpretable, as the prompt $s$ is expressed in  natural language. 

    This connection reveals that prompt optimization within the natural language space offers a form of interpretable learning that simultaneously achieves both expressiveness and interpretability. The key to bridging this gap lies in leveraging LLMs to handle the non-interpretable processing of raw signals into high-level concepts, much like how neurons in the human brain transform signals into information. This allows learning to occur within an interpretable space.

    \paragraph{Limitation of (discrete) prompt optimization}
    However, existing prompt optimization algorithms are insufficient for interpretable learning for several reasons: firstly most methods focus on ``rewriting'' prompts to enhance performance~\citep{apo, long_prompt}, which might not help in extracting interpretable knowledge from data. Additionally, while recent developments show some capabilities in correcting prompts using error examples~\citep{apo, promptagent}, they still struggle with complex decision rules, such as conditional branching for classification tasks. These rules, often applicable to only a subset of samples, are difficult to recover when considering the whole training set. Our experiments indicate that direct application of current methods fails to effectively address complex decision rules. These challenges motivate the proposed LSP framework that combines prompt optimization with symbolic programs.

    \vspace{-5pt}
    \subsubsection{Domain-Specific Language of LSPs}
    \label{sec:method.lsp.dsl}
    \vspace{-5pt}
    Compared with traditional NSPs that require manually designing a comprehensive DSL, LLM's ability to represent a wide range of functions via different prompting, we can significantly streamline the grammar required to build expressive and interpretable models.
    Specifically, for predictive models, we can build powerful LSPs from a minimalist DSL with only three components: the input, conditional branching, and LLM module:in
    \begin{align}
        \alpha &::= x \mid \texttt{switch}(\{\alpha == y_i: \alpha_i \}_{i=1}^{k}) \mid \texttt{LLM} (x, s).
        \label{eq:cfg.lsp}
    \end{align}
    Here,
    \textbf{input $x$} represents the input data (text, image, etc);
    the \textbf{conditional branching $\texttt{switch}(\{y_i: \alpha_i \}_{i=1}^{k})$} forms the backbone of the program structure.
    Each switch can be viewed as a node in a decision tree tree with $k$ branches. 
    It will branch to $\alpha_i$
    if the sub-program $\alpha$ predicts $y_i$. 
    \textbf{The LLM Module $\texttt{LLM}(x, s)$} serves as the inference engines. It means to prompting LLM to make a prediction on input $x$ under the instruction $s$.

    Figure~\ref{fig:inference} shows an example LSP generated from above DSL.
    Given a test query, we traverse the tree-structured program in a top-down manner, assigning data to specific child node based on the parent node's predictions, until the leaf node is reached and the final response is returned.

\vspace{-5pt}
    \subsubsection{Learning algorithm}
    \label{sec:method.lsp.algo}
    \vspace{-5pt}

\begin{figure}[t]
\vspace{-20pt}
    \centering
    \includegraphics[width=1.0\textwidth, trim={50mm 65mm 25mm 55mm}, clip]{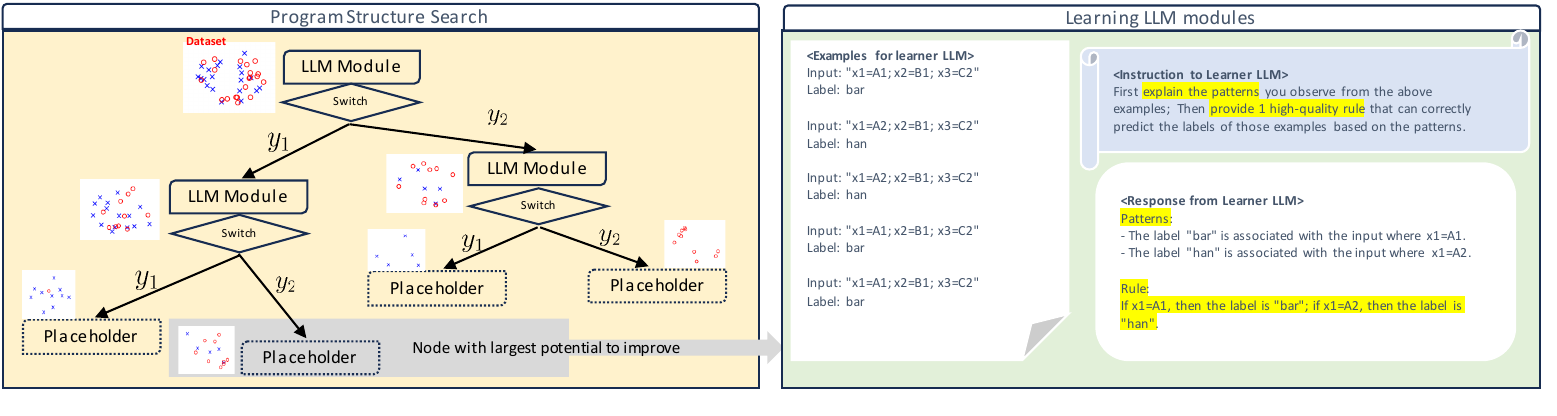}
    \caption{
        \small
        \textbf{Learning Algorithm for LSPs.} The learning algorithm for LSPs contains two parts: \textbf{(1) program structure search (Left):} This process is akin to constructing a traditional decision tree. Starting from the root, the algorithm traverses down the tree, iteratively splitting the training dataset based on the current node's predictions and expanding the leaf node with the highest prediction errors. \textbf{(2) LLM module optimization (Right):} Here, a learner LLM is instructed to summarize rules based on the observed data at its node.
    }
    \vspace{-10pt}
    \label{fig:training}
\end{figure}

    \paragraph{Overview}
    The tree search framework commonly used in NSPs also applies to LSPs~\citep{nsp, near}.
    Moreover, the simplicity of our DSL allows the search process to be further arranged in a highly intuitive manner akin to constructing decision trees.
    As illustrated in Figure~\ref{fig:training}, the process begins at the root node with an empty program and the entire training set. A $\texttt{switch}$ operator combined with an $\texttt{LLM(x, s)}$ module is then added.
    This combination essentially directs the program's flow based on the module's predictions.
    Once the LLM module is trained, we expand its child nodes and repeat the whole process.
    This divide-and-conquer strategy benefits the search process by simplifying each LLM module's task to fitting only a subset of the data.
    
    \paragraph{Learning LLM-modules by summarizing predictive rules}
    In LSPs, each LLM module is responsible for decision-making on its designated data subset. 
    For traditional NSPs, the neural modules are optimized using empirical risk minimization;
    For LSPs, training LLM modules essentially transforms into deriving rules from observed data, as established in Section~\ref{sec:method.lsp.theory}.
    While this can be achieved via generic prompt optimization techniques, we adopt a more direct approach utilizing the LLM’s robust summarization capabilities~\citep{summary1, summary2, summary3, summary4}, asking the model to summarize rules from observed data patterns.
    Note that LLM serves as both the inference engine and the learner, as depicted in Figure~\ref{fig:training} (Right).

    \paragraph{Node selection}
    Top-down tree search algorithms generally use a node scoring function to determine the next node to expand, ideally prioritizing nodes with the greatest potential for program improvement.
    Given that nodes with more frequent errors likely have more room for improvement, we use error count as our scoring function.
    This metric, accounting for both the error rate and the size of the data subset each node handles, provides a simple yet empirically effective approach.
    Section~\ref{sec.ablate} presents empirical evidence supporting the efficacy and robustness of this metric.

    \paragraph{Complete algorithm}
    The above outlines the learning process for expanding a single program.
    In the full search pipeline, we further incorporate beam search~\citep{apo} and batch sampling to prevent the search from getting stuck in local minima, as summarized in Algorithm~\ref{algo:search_algorithm} (Appendix).


\vspace{-5pt}
\section{IL-Bench Interpretable-Learning Benchmark}
\label{sec.method.benchmark}
\vspace{-5pt}
The goal for interpretable learning is for the model to acquire knowledge transferable to humans for classification tasks that are NOT zero-shot solvable. A task is zero-shot solvable if the LLM can predict accurately based solely on the class name (e.g. basic dog-cat classification). To evaluate interpretable learning methods, we need classification tasks with classes unseen during LLM pretraining, requiring the model to learn additional knowledge for successful classification.
This is challenging given the extensive pretraining on Internet data.
Therefore, we introduce Interpretable-Learning Bench, a novel benchmark comprising multiple challenging tasks, and even advanced LLMs like GPT need to learn substantial additional knowledge to solve these tasks. This benchmark provides a valuable resource for evaluating future interpretable learning and autoprompting methods. We will explain the construction of IL-Bench next, and leave the detailed summaries to Appendix Table~\ref{tab:il_bench_overview}.

\paragraph{Synthetic Tasks}
\label{sec.method.benchmark.synthetic}
Prior work in symbolic learning often uses synthetic datasets to evaluate methodologies due to known oracle rules, making it easy to observe model performance. Inspired by this, we develop synthetic datasets with known predictive rules in our benchmark. These datasets employ symbols to represent variables and values, making them context-agnostic and requiring the model to rely on abstract reasoning. Our tasks include decisions generated by decision trees of varying complexity, allowing us to assess model behavior as rule complexity increases.

\paragraph{Textual Classification Tasks: from image to text dataset}
\label{sec.method.benchmark.caption}
To evaluate the model's proficiency in complex scenarios, Fine-Grained Visual Classification (FGVC) tasks~\citep{fgvc, cub, cars, flowers, nabirds} serve as an excellent testbed.
FGVC focuses on distinguishing objects within narrowly defined categories, where the differences between examples from each class are often subtle.
We selected specific image subsets from FGVC databases, covering area of various bird species~\citep{cub}.
For evaluating pure-text LLMs, we convert them to text datasets via captioning.
To ensure the captions cover the intricate differences between objects, we introduce \textbf{the Concatenated Contrastive Captioning (C3) technique}, which generates detailed captions by contrasting a target image with reference images from different classes, merging them into a comprehensive description for text-only LLM evaluation.

\paragraph{Visual classification Tasks: distinguishing novel visual concepts}
\label{sec.method.benchmark.vision}
We also collect a new suit of datasets from Palworld, a Pokemon-style game containing various species of creatures (Examples in Table~\ref{tab:il_bench_overview}).
Since the game is released after most existing LLM's knowledge cut-off date, the model will need to rely solely on the knowledge extracted from the dataset to perform the prediction.

\section{Related Work}
\label{sec.related}

\paragraph{Interpretable machine learning}
Although neural networks are immensely expressive, they provide no insights into its internal decision making mechanism.
In the quest of making model predictions interpretable, research has broadly categorized methods into two main types: post-hoc and intrinsic.
Post-hoc methods provide insights into how a pretrained model behaves, usually by highlighting important features used for decision making~\citep{zintgraf2017visualizing, petsiuk2018rise, dabkowski2017real, shrikumar2017learning, sundararajan2017axiomatic,ancona2017towards} or provide counterfactual explanations~\citep{dhurandhar2018explanations,Hendricks2018GroundingVE,vanderwaa2018,goyal2019counterfactual,hsieh2021evaluations}. Beyond attribution in the feature space, some methods can also be generalized to the space of higher level concepts~\citep{kim2018interpretability,bai2022concept}. However, all these methods aim to highlight important features while not being able to recover the entire decision making process of neural networks.

On the other hand, intrinsic methods integrate interpretability directly into the model's architecture, making them naturally interpretable by design.
Traditional Methods include Decision Trees~\citep{xgboost} and Generalized Additive Models (GAMs)~\citep{gam} offer strong interpretability, yet often not expressive enough. Concept bottleneck model adds a hidden layer in neural network, where neurons represent some predefined concepts to gain interpretability~\citep{koh2020concept,losch2019interpretability,yuksekgonul2022post,oikarinen2023label}. While this approach facilitates attribution of concepts, it does not provide a comprehensive decision rule.
Neurosymbolic Programming (NSP)~\citep{nsp, near, dpad, prototree} represents an innovative blend, combining deep learning's data handling capabilities with symbolic reasoning to foster both performance and transparency.
Despite early promises, NSP suffers from an inherit trade-off between expressiveness (more NN modules) and interpretability (more symbolic modules).
Moreover, they are often expensive to train due to co-optimization of program architecture and parameters of the NN modules~\citep{near, dpad}.

\paragraph{Prompt Optimization}
The essence of utilizing a generative language model lies in crafting effective prompts.
Recent advancements have aimed to automate this process, reducing the need for human effort through prompt optimization~\citep{autoprompt, ape}.
While pioneering efforts were mainly directed towards various discrete optimization algorithms~\citep{autoprompt, rlprompt, tempera}, it has been noted that advanced LLMs can revise prompts similarly to human engineers~\citep{ape, apo}.
Since these initial efforts, a significant body of research has emerged, exploring various search algorithms including Monte Carlo Sampling~\citep{ape}, beam search~\citep{apo}, evolutionary search~\citep{opro, promptbreeder, gps, genetic, long_prompt}, and tree search~\citep{promptagent}.
However, existing methods often treat the prompt as a single entity without explicit structure.
From this perspective, prompt optimization methods can be seen as simplified instances of LSPs, where the program consists solely of one LLM module. While this simplification has shown promising results, as task complexity increases, the explicit structuring within LSPs allows them to encode knowledge from data. This provides substantial advantages over conventional prompt optimization methods.

\vspace{-5pt}
\section{Experimental Results}
\label{sec:exp}
\vspace{-5pt}
We adopt a comprehensive approach to extensively evaluate the effectiveness of LSPs against various baselines under different settings.
Our empirical study is designed to validate the benefits of LSPs over alternative methods by addressing the following research questions:
\begin{itemize}[noitemsep, topsep=0pt, parsep=0pt, partopsep=0pt, leftmargin=*]
    \item \textbf{Q1: How does LSP compare against traditional NSPs in expressiveness and interpretability?} We assess this through both quantitative and qualitative evaluations (human studies). (Section~\ref{sec:exp.xai})
    \item \textbf{Q2: Does LSP generalize better than traditional NSPs under domain shifts?} This question is explored in detail in Section~\ref{sec:exp.xai}.
    \item \textbf{Q3: Is the incorporation of explicit structures beneficial to LSPs?} We compare the structured LSP with vanilla prompt optimization, which exemplifies a special case of LSP with a single LLM module. (Section~\ref{sec:exp.po})
    \item \textbf{Q4: How effective are different LLMs in implementing LSP?} We conduct cross-model experiments to evaluate the performance of various LLMs as the computational backbone for learning and inference in LSP. (Section~\ref{sec:exp.cross_model})
\end{itemize}

\subsection{General settings}
\label{sec:exp.settings}
\paragraph{Evaluation}
For language tasks, we test popular LLMs, including GPT-3.5 (\texttt{turbo-1104})~\citep{gpt3}, GPT-4 (\texttt{1106-preview})~\citep{gpt4}, and Gemini-M (\texttt{1.0-pro})~\citep{gemini}.
For vision tasks, GPT-4V (\texttt{gpt-4-1106-vision-preview}) and Gemini-Vision (\texttt{1.0-pro-vision}) are utilized.
All experiments are repeated with 3 seeds.

\paragraph{Implementation details of LSP}
Our default model of choice is GPT-3.5 for language tasks and Gemini-Vision for vision tasks, but also examine cross-(M)LLM performance in Appendix.
All LLM modules are initialized with an empty instruction ``none''.
More detailed hyperparameters can be found in Appendix, which is kept fixed throughout the experiments.

\subsection{Comparison with traditional interpretable learning methods}
\label{sec:exp.xai}
\begin{table}
    \caption{\small \textbf{Classification accuracy comparison with XAI methods on IL-Bench-Vision.} Here, all numbers for LSP are obtained with Gemini-Vision as the learner and inference LLM, except for LSP (GPT-4V) which uses the larger GPT-4V as the learner; Decision Tree, operating directly on pixel data, lacks human interpretability. Key findings include: (1) Our method outperforms XAI baselines with an average accuracy of 95.67\%, which is over 10\% higher than the nearest competitor. (2) The program generated by LSP also demonstrates superior transferability to human raters, as they are able to reproduce the predictions following rules learned by LSP.}
    \label{tab:main_exp_nsp}
    \centering
    \begin{threeparttable}
    \resizebox{1\textwidth}{!}{
    \begin{tabular}{cc|cccccccc}
        \toprule
        
        \midrule
        IL-Bench-Vision &
        & \multicolumn{8}{c}{Palworld} \\
        \midrule
        MLLM
        & Method
        & Mean
        & Fire-1 & Fire-2
        & Dragon-1 & Dragon-2
        & Electric-1 & Electric-2
        & Water-1 \\
        \midrule
        
        \multirow{6}{*}{Gemini-M}
            & Decision Tree~\citep{xgboost}
            & 68.20
            & 91.11 $\pm$ 12.57 & 32.00 $\pm$ 9.80
            & 68.33 $\pm$ 10.27 & 48.33 $\pm$ 20.95
            & 82.67 $\pm$ 6.80 & 65.33 $\pm$ 13.60
            & 66.67 $\pm$ 8.50 \\
            \cmidrule(r){2-10}
            & ProtoTree~\citep{prototree}
            & 84.33
            & \bf{100.00 $\pm$ 0.00} & 62.67 $\pm$ 12.36
            & \bf{98.33 $\pm$ 2.36} & 85.00 $\pm$ 4.08
            & \bf{100.00 $\pm$ 0.00} & 82.67 $\pm$ 9.98
            & 61.67 $\pm$ 25.93 \\
            \cmidrule(r){2-10}
            & LSP
            & 84.76
            & 93.33 $\pm$ 6.67 & 76.00 $\pm$ 4.00
            & 85.00 $\pm$ 0.00 & \bf{97.50 $\pm$ 2.50}
            & 86.00 $\pm$ 10.00 & 68.00 $\pm$ 4.00
            & 87.50 $\pm$ 2.50 \\
            \cmidrule(r){2-10}
            & LSP (GPT-4V)
            & \bf{95.67}
            & 96.67 $\pm$ 3.33 & \bf{90.00 $\pm$ 6.00}
            & 90.00 $\pm$ 10.00 & \bf{97.50 $\pm$ 2.50}
            & \bf{100.00 $\pm$ 0.00} & \bf{98.00 $\pm$ 2.00}
            & \bf{97.50 $\pm$ 2.50} \\
        \midrule

        \multirow{2}{*}{Human Rater}
            & ProtoTree~\citep{prototree}
            & 72.74
            & 83.33 $\pm$ 16.67 & 50.0 $\pm$ 10.0
            & \bf 100.0 $\pm$ 0.0 & 75.0 $\pm$ 0.0
            & 83.33 $\pm$ 16.67 & 80.0 $\pm$ 0.0
            & 37.5 $\pm$ 12.5 \\
            \cmidrule(r){2-10}
            & LSP (GPT-4V) 
            & \bf{90.36}
            & \bf 100.00 $\pm$ 0.00 & \bf 70.00 $\pm$ 10.00
            & \bf 100.00 $\pm$ 0.00 & \bf 87.5 $\pm$ 12.5
            & \bf 100.00 $\pm$ 0.00 & \bf 100.00 $\pm$ 0.00
            & \bf 75.00 $\pm$ 25.00 \\
        \midrule

        \bottomrule

    \end{tabular}
    }
    \end{threeparttable}
    \vspace{-3mm}
\end{table}

\begin{wrapfigure}{r}{0.4\textwidth} 
    \vspace{-8mm}
    \centering
    \includegraphics[width=0.4\textwidth]{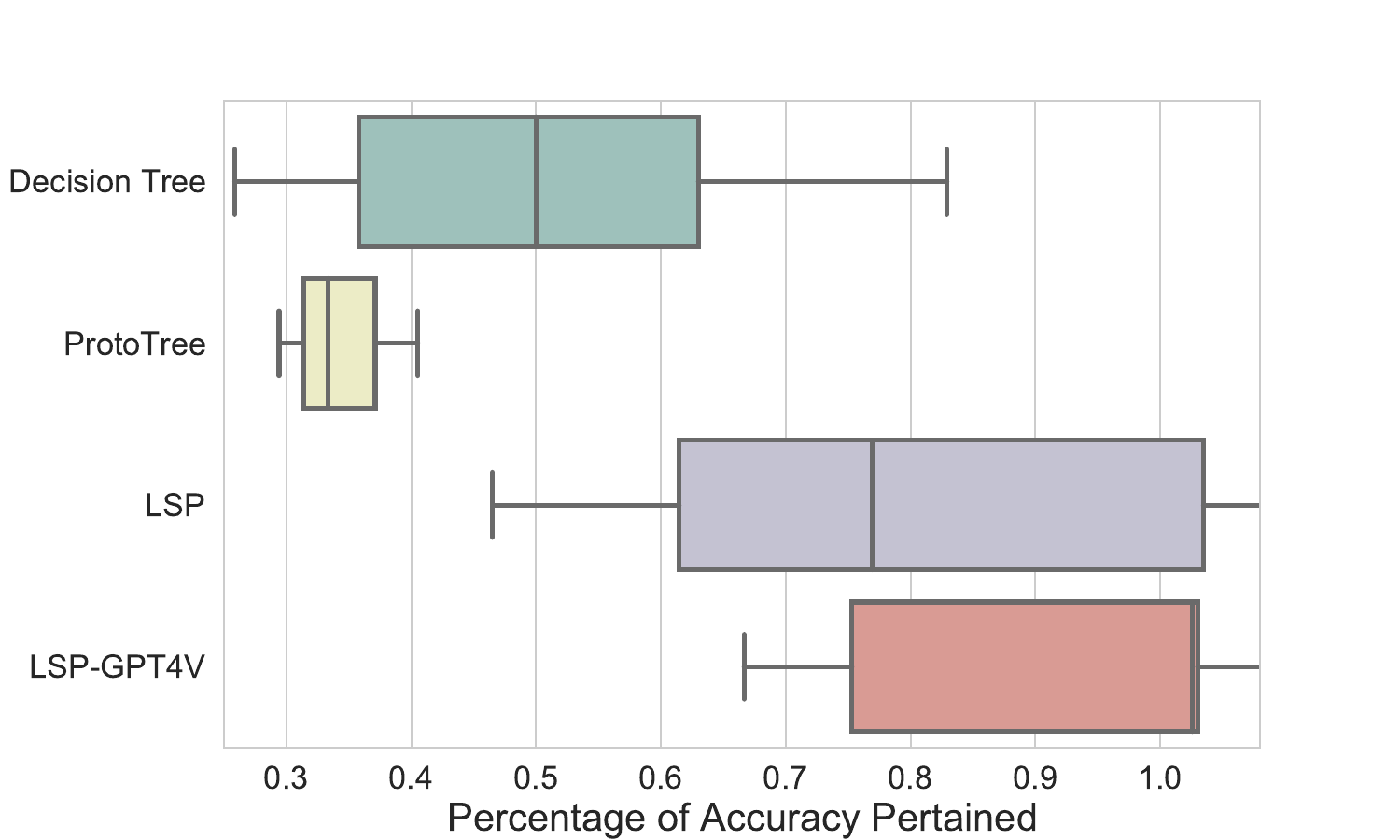}
    \caption{
        \small
        \textbf{Accuracy retention rate on Out-Of-Distribution variants of IL-Bench-Vision testsets.} We compute the ratio of test accuracy evaluated on OOD datasets to the original test accuracy. LSP shows strong transferability to OOD data. Notably, the version using GPT-4V as the learner retains 90-100\% of the original test accuracy.
    }
    \label{fig:ood_result}
    \vspace{-6mm}
\end{wrapfigure}
We compare LSP with two established models - ProtoTree~\citep{prototree} and Decision Tree~\citep{xgboost} - both organize prediction process in tree-structured formats.
Among existing NSP methods, the closest to ours is ProtoTree - a highly interpretable NSP that learns a discrete binary tree end-to-end, where each node stores an image patch ("prototype") and the edges determine whether the prototype exists within the query image.
Note that ProtoTree does not rely on an explicit DSL - we could not compare with methods based on explicit DSL since they require domain experts to design those operation, while our goal is to automate the whole process.
Since ProtoTree only implements image tasks, this comparison also focus on the vision tasks in IL-Bench.

\paragraph{Expressiveness}
The expressiveness of the learned programs is evaluated in Table~\ref{tab:main_exp_nsp}.
LSP (GPT4V) outperforms ProtoTree with an average accuracy of 95.67\%, which is over 10\% gain.
Considering that GPT/Gemini has never observed the images in our datasets before (curated after their knowledge cutoff), this result suggests LSP is capable of formulating effective predictive rules from previously unseen examples.

\paragraph{Interpretability}
We measure the interpretability of LSPs and NSPs by having human raters make predictions based on visualizations of the learned programs (See Appendix for evaluation protocols).
This process essentially "transfers" knowledge from models back to human.
Notably, many XAI methods fall short of achieving this level of interpretability, with ProtoTree being a rare exception.
As summarized in Table~\ref{tab:main_exp_nsp}, the program generated by LSP also demonstrates stronger transferability to human raters, as they are able to largely reproduce the predictions following rules learned by LSP.

\paragraph{Generalization under Domain Shift}
In contrast to traditional NSP models that rely on parametric memory, LSP utilizes language instructions to encode knowledge.
This strategy significantly enhances robustness against variations in visual attributes (domain shifts).
To verify this advantage, we examine the transferability of the learned programs to Out-of-Distribution (OOD) data, constructed using GPT-4V (See Appendix for details)
As shown in Figure~\ref{fig:ood_result}, LSP demonstrates exceptional resilience to domain shifts, compared with ProtoTree.

\subsection{Comparison with prompt optimization}
\label{sec:exp.po}
\begin{table}
  \caption{\small \textbf{Classification accuracy comparison with Prompt Optimization methods on IL-Bench-Language.} Key findings include: (1) LSP achieves $\sim 5\%$ accuracy gain over PromptAgent, the previous state-of-the-art. (2) Across synthetic Decision Tree datasets categorized by increasing complexity of oracle decision rules (Easy, Medium, Hard), LSP consistently outperforms other methods in maintaining high accuracy levels, demonstrating its superior ability to reverse-engineer complex rules from observed data.}
  \label{tab:main_exp_text}
  \centering
  \resizebox{1\textwidth}{!}{
  \begin{tabular}{cc|c|ccc|ccccccc}
    \toprule
    
    \midrule
    Text Benchmark & 
    & \multicolumn{5}{c}{Synthetic}
    & \multicolumn{4}{c}{Caption} \\
    \midrule
    LLM
    & Method
    & Mean
    & DT-Easy & DT-Medium & DT-Hard
    & Waxwing & Waterthrush & Jaeger & Albatross & Blackbird & Swallow \\
    \midrule
    
    \multirow{6}{*}{GPT-3.5}
        & APE~\citep{ape}
        & 67.62
        & 99.67 $\pm$ 0.47 & 86.67 $\pm$ 12.47 & 87.00 $\pm$ 8.57
        & 46.11 $\pm$ 4.37 & 43.89 $\pm$ 3.14 & 65.56 $\pm$ 2.83 & 47.41 $\pm$ 2.28 & 78.06 $\pm$ 1.04 & 54.17 $\pm$ 1.18 \\
        \cmidrule(r){2-12}
        & OPRO~\citep{opro}
        & 55.48
        & 50.00 $\pm$ 1.08 & 50.17 $\pm$ 3.06 & 30.33 $\pm$ 2.62
        & 57.22 $\pm$ 2.08 & 57.22 $\pm$ 4.16 & 76.67 $\pm$ 4.71 & 40.37 $\pm$ 3.43 & 78.06 $\pm$ 2.83 & 55.28 $\pm$ 1.04 \\
        \cmidrule(r){2-12}
        & APO~\citep{apo}
        & 70.67
        & \bf{100.00 $\pm$ 0.00} & 96.67 $\pm$ 4.71 & 77.83 $\pm$ 11.90
        & 56.11 $\pm$ 4.78 & 48.89 $\pm$ 4.16 & 70.00 $\pm$ 5.93 & 54.07 $\pm$ 9.70 & 74.17 $\pm$ 2.97 & 58.33 $\pm$ 1.36 \\
        \cmidrule(r){2-12}
        & PromptAgent~\citep{promptagent}
        & 72.40
        & 97.67 $\pm$ 3.30 & 88.50 $\pm$ 8.44 & 64.33 $\pm$ 20.27
        & 60.56 $\pm$ 4.78 & 56.67 $\pm$ 6.24 & 75.00 $\pm$ 3.60 & \bf{74.44 $\pm$ 6.54} & 74.17 $\pm$ 1.36 & 57.22 $\pm$ 0.79 \\
        \cmidrule(r){2-12}
        & LSP (Ours)
        & \bf{77.29}
        & 99.25 $\pm$ 0.75 & \bf{98.50 $\pm$ 0.00} & \bf{89.75 $\pm$ 1.25}
        & \bf{65.83 $\pm$ 4.17} & \bf{62.50 $\pm$ 0.83} & \bf{80.00 $\pm$ 1.67} & 61.11 $\pm$ 1.11 & \bf{78.75 $\pm$ 0.42} & \bf{62.92 $\pm$ 0.42} \\
    \midrule
    
    \bottomrule

  \end{tabular}
  }
  \vspace{-3mm}
\end{table}

\begin{table}
  \caption{\small \textbf{Classification accuracy comparison with Prompt Optimization methods on IL-Bench-Vision.} LSP achieves an average accuracy of 84.47\%, which is $\sim 20\%$ higher than the 2nd best method (APE).}

  \label{tab:main_exp_vision}
  \centering
  \resizebox{1\textwidth}{!}{
  \begin{tabular}{cc|cccccccc}
    \toprule
    
    \midrule
    Vision Benchmark &
    & \multicolumn{7}{c}{Palworld} \\
    \midrule
    MLLM
    & Method
    & Mean
    & Fire-1 & Fire-2
    & Dragon-1 & Dragon-2
    & Electric-1 & Electric-2
    & Water-1 \\
    \midrule
    
    \multirow{6}{*}{Gemini-M}
        & APE~\citep{ape}
        & 64.38
        & 76.67 $\pm$ 3.33 & 56.00 $\pm$ 0.00 
        & 59.17 $\pm$ 5.83 & 35.00 $\pm$ 5.00 
        & 57.33 $\pm$ 10.67 & \bf 82.00 $\pm$ 2.00 
        & 82.50 $\pm$ 12.50 \\
        \cmidrule(r){2-10}
        & OPRO~\citep{opro}
        & 49.26
        & 86.67 $\pm$ 0.00 & 32.00 $\pm$ 4.00 
        & 56.67 $\pm$ 3.33 & 50.00 $\pm$ 5.00 
        & 48.00 $\pm$ 28.00 & 49.00 $\pm$ 19.00 
        & 27.50 $\pm$ 2.50 \\
        \cmidrule(r){2-10}
        & APO~\citep{apo}
        & 58.45
        & 76.67 $\pm$ 23.33 & 42.00 $\pm$ 14.00 
        & 59.17 $\pm$ 5.83 & 70.00 $\pm$ 5.00 
        & 39.33 $\pm$ 7.33 & 67.50 $\pm$ 12.50 
        & 52.50 $\pm$ 17.50 \\
        \cmidrule(r){2-10}
        & PromptAgent~\citep{promptagent}
        & 54.74
        & 66.67 $\pm$ 6.67 & 44.00 $\pm$ 0.00 
        & 49.17 $\pm$ 15.83 & 52.50 $\pm$ 2.50 
        & 55.33 $\pm$ 8.67 & 42.50 $\pm$ 2.50 
        & 70.00 $\pm$ 0.00 \\
        \cmidrule(r){2-10}
        & LSP (Ours)
        & \bf 84.47
        & \bf 93.33 $\pm$ 6.67 & \bf 76.00 $\pm$ 4.00 
        & \bf 85.00 $\pm$ 0.00 & \bf 97.50 $\pm$ 2.50 
        & \bf 86.00 $\pm$ 10.00 & 68.00 $\pm$ 4.00 
        & \bf 87.50 $\pm$ 2.50 \\
    \midrule
    
    \bottomrule

  \end{tabular}
  }
  \vspace{-3mm}
\end{table}

Since there exists a variety of PO method that primarily differ in the search algorithm, we select one most representative method from each major category: Monte Carlo sampling (APE)~\citep{ape}, evolutionary search (ORPO)~\citep{opro}, beam search (APO)~\citep{apo}, and tree search (PromptAgent)~\citep{promptagent}.
Since the main bottleneck for PO methods is the candidate evaluation, we follow existing works and set the same maximum number of candidate proposals for all methods (100 candidates).

\paragraph{Results}
The empirical results indicate that incorporating explicit structures significantly enhances performance of the programs on predictive tasks:
LSP consistently outperforms all vanilla prompt optimization methods, with a considerable margin of $20.09\%$ and $4.89\%$ over the 2nd best methods on vision and language tasks respectively.
The advantages of integrating structured learning are twofold:
(1) It simplifies the learning process: LSP benefits from a divide-and-conquer approach where each LLM-module node focuses solely on extracting predictive rules for a specific subset of the data.
(2) It streamlines the inference process: We observe that LLMs tend to exhibit hallucination as the complexity of the instructions increases (e.g., multiple conditional clauses.
In contrast, LSP mitigates this issue by ensuring that each LLM module contains simpler, more manageable instructions.

\begin{figure}[t]
    \centering
    \begin{subfigure}{0.24\textwidth}
        \includegraphics[height=1.4in, width=\linewidth]{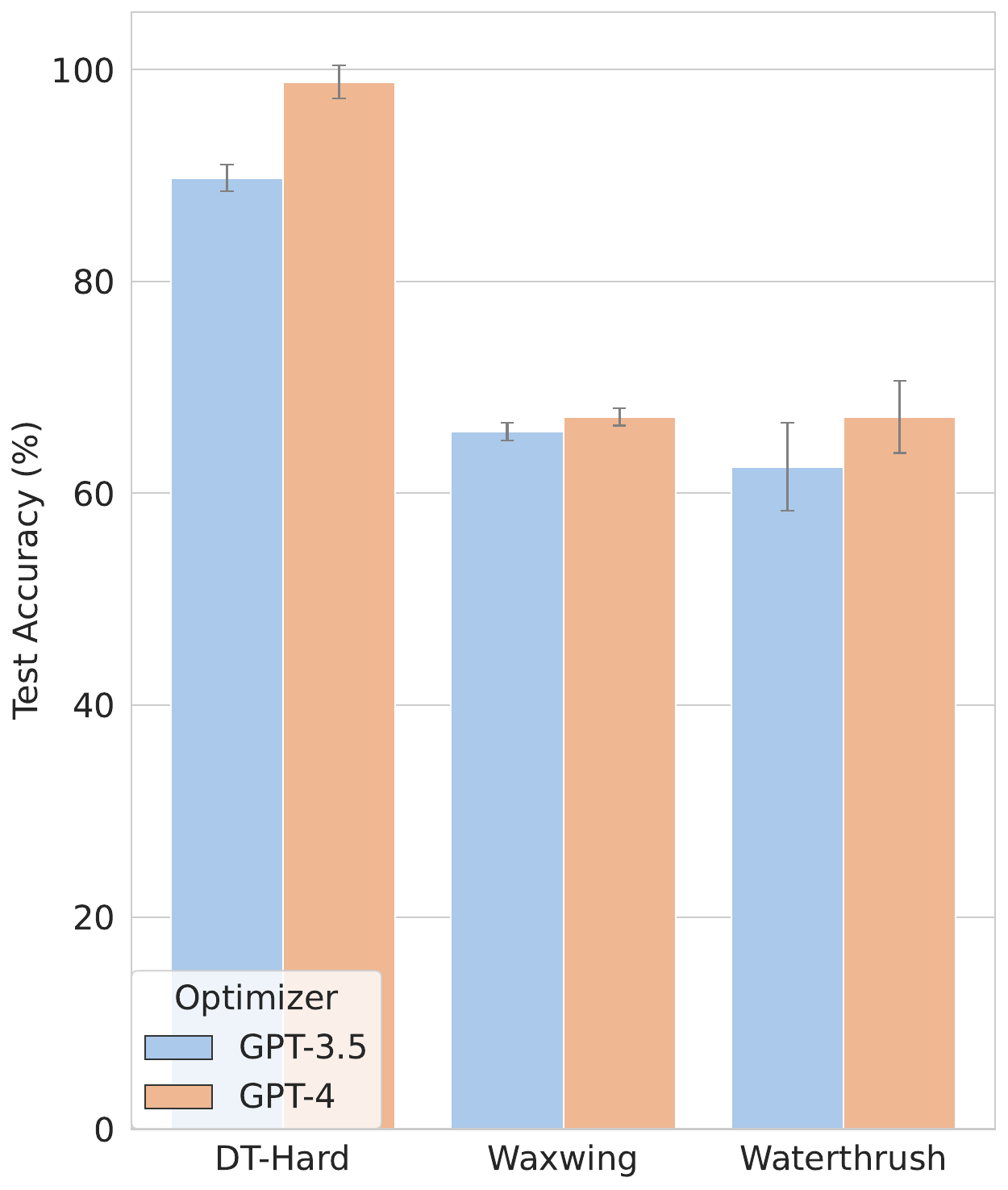}
        \vspace{-10pt} 
        \caption{Language Tasks}
        \label{fig:optimizer.text}
    \end{subfigure}
    \hfill 
    \begin{subfigure}{0.24\textwidth}
        \includegraphics[height=1.4in, width=\linewidth]{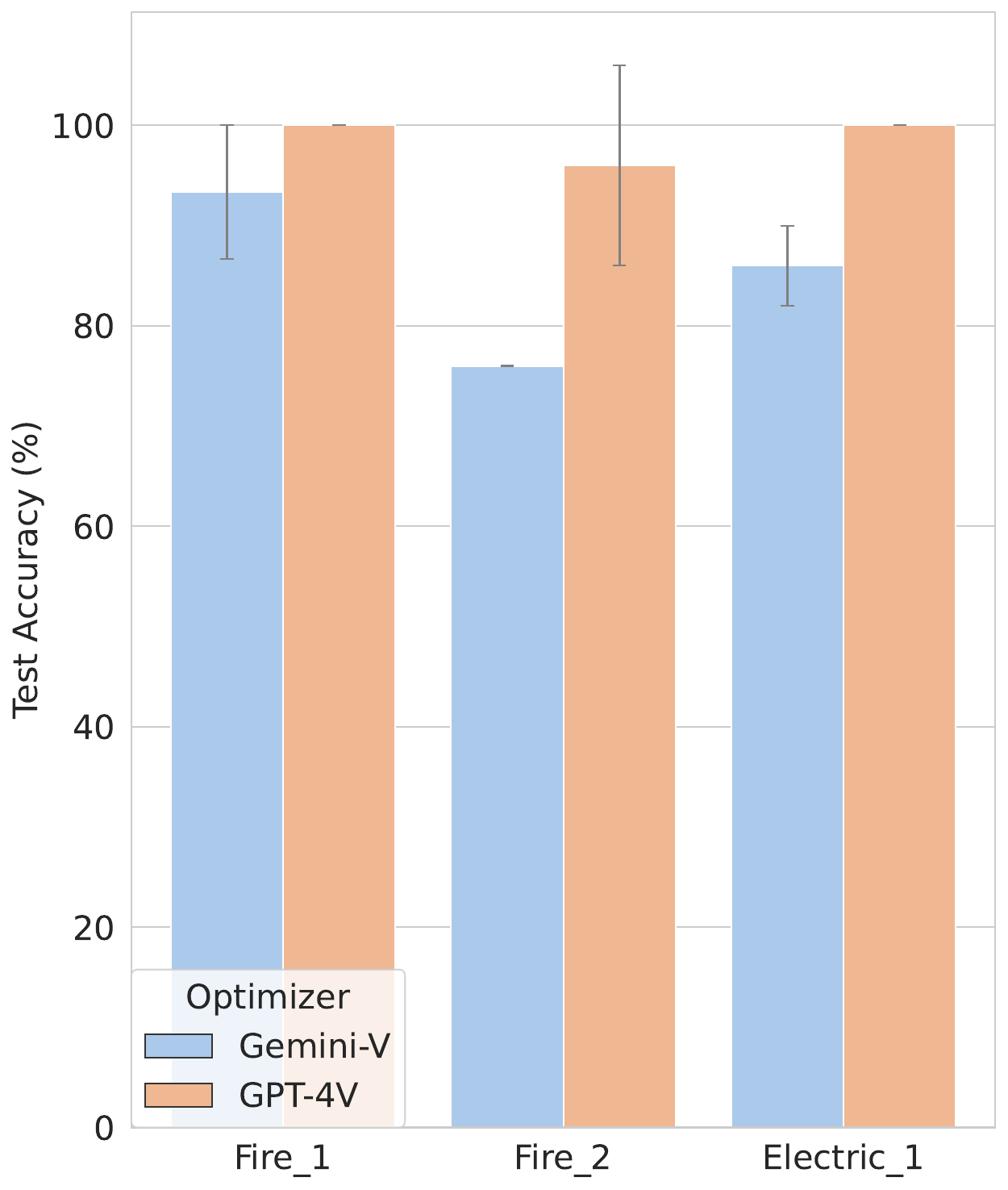}
        \vspace{-10pt} 
        \caption{Vision Tasks}
        \label{fig:optimizer.vision}
    \end{subfigure}
    \hfill 
    \begin{subfigure}{0.24\textwidth}
        \includegraphics[height=1.4in, width=\linewidth]{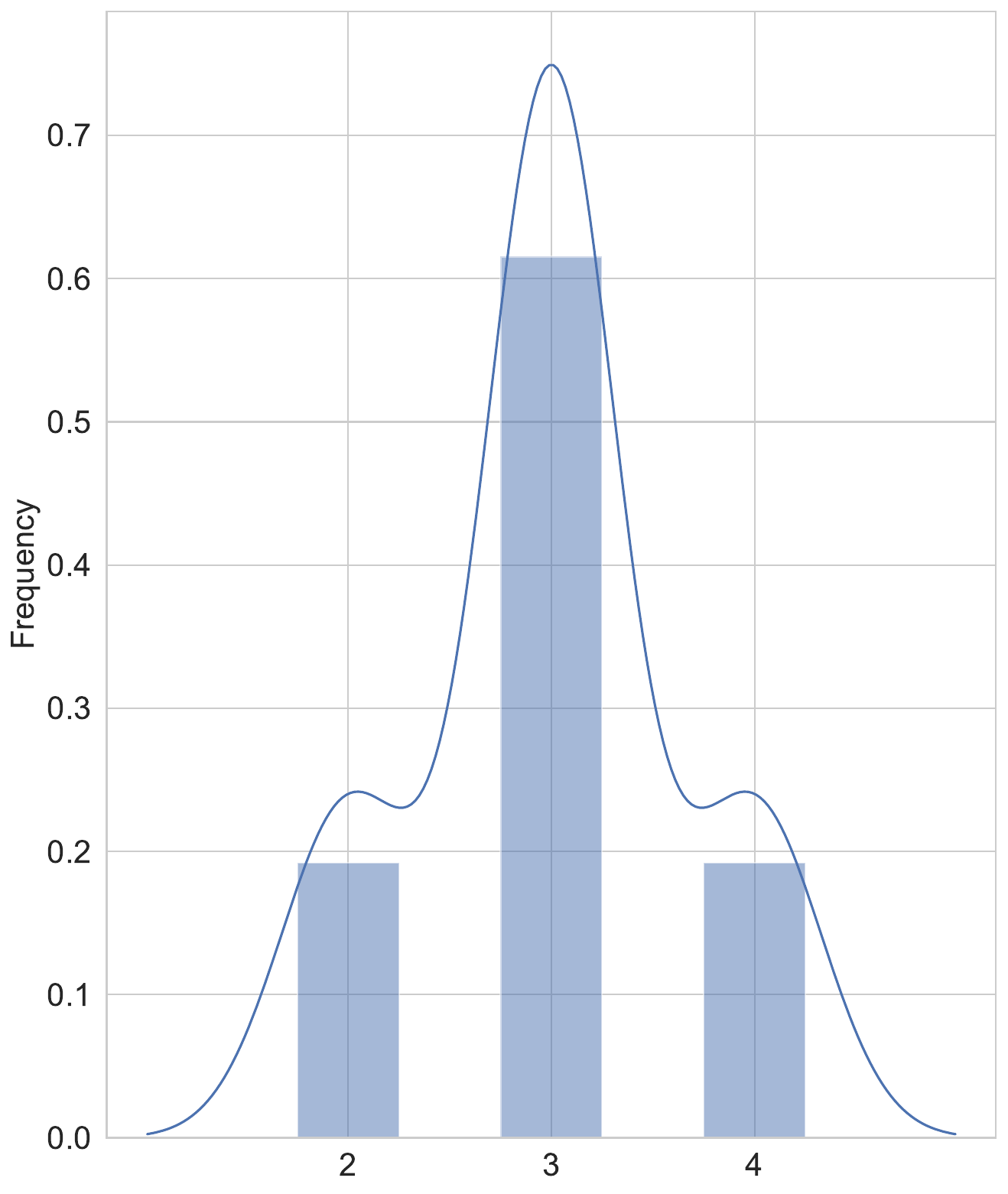}
        \vspace{-10pt} 
        \caption{Program Depth}
        \label{fig:tree_stats.depth}
    \end{subfigure}
    \hfill 
    \begin{subfigure}{0.24\textwidth}
        \includegraphics[height=1.4in, width=\linewidth]{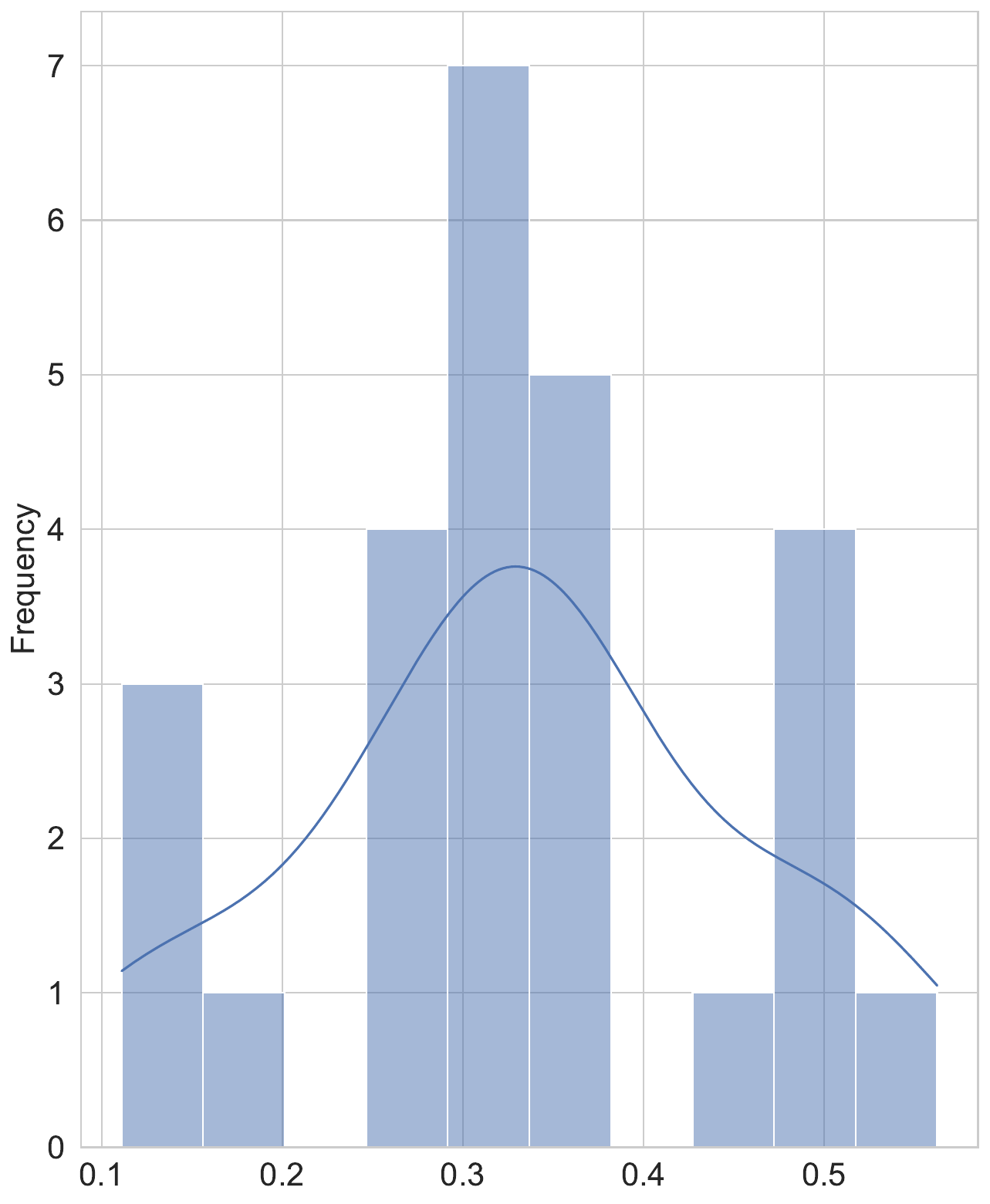}
        \vspace{-10pt} 
        \caption{Program Sparsity}
        \label{fig:tree_stats.sparsity}
    \end{subfigure}
    \caption{
        \small
        \textbf{(a, b): Stronger LLMs as better LSP learners.} In these experiments, we keep the inference LLM fixed (GPT-3.5 for text and Gemini-V for images) while swapping the learner LLM with GPT-4. With its larger parameter count, GPT-4 consistently achieves better performance in learning LSPs.
        \textbf{(c, d): Statistics of discovered programs.} Averaged from the IL-Bench-Language tasks, the resulting LSPs are generally shallow and sparse, indicating that the final prediction can be reached within only a few steps.
    }
    \label{fig:optimizer}
\end{figure}

\section{Ablation Study}
\label{sec.ablate}

\paragraph{Convergence of LLM-Symbolic Program LSP}
\begin{figure}[t]
\vspace{-10pt}
    \centering
    \begin{subfigure}{0.24\textwidth}
        \includegraphics[width=\linewidth]{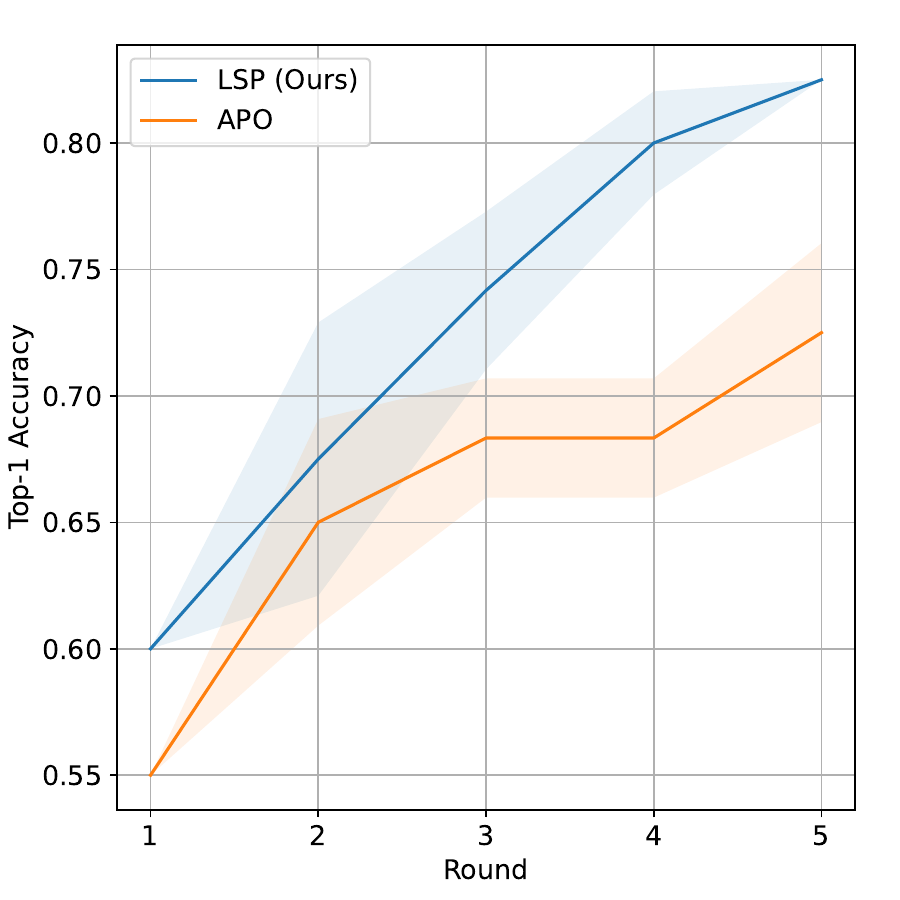} 
        \caption{CUB-Waxwing}
        \label{fig:sub1}
    \end{subfigure}
    \hfill 
    \begin{subfigure}{0.24\textwidth}
        \includegraphics[width=\linewidth]{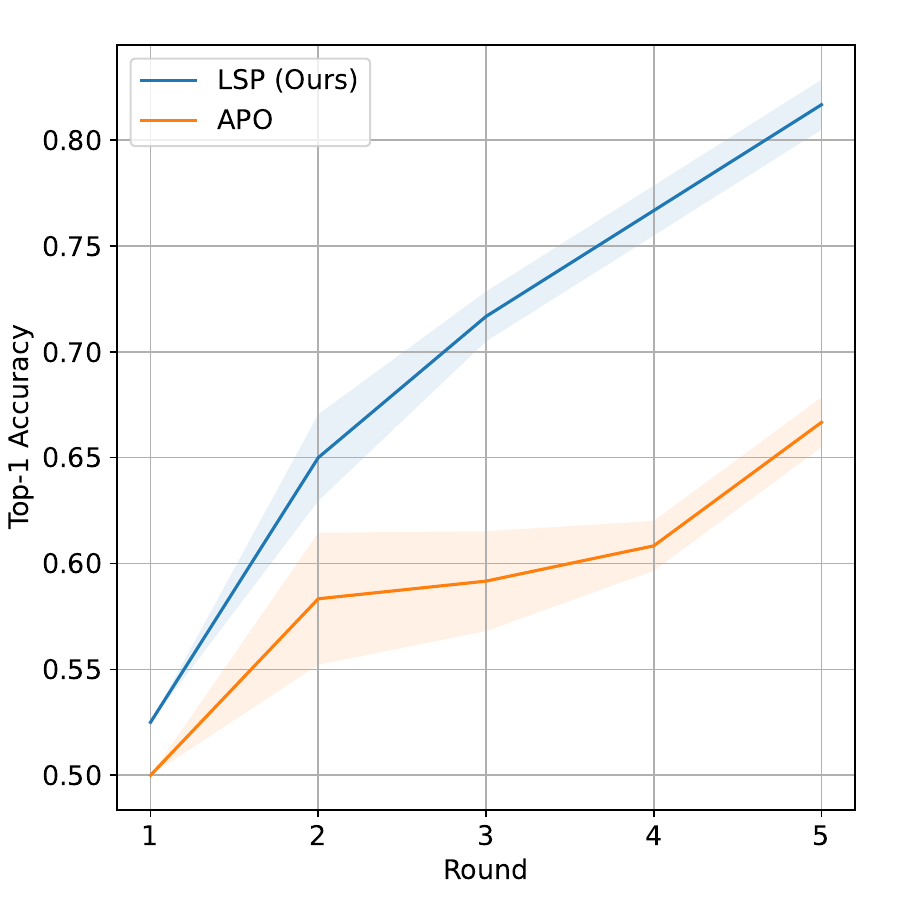}
        \caption{CUB-Waterthrush}
        \label{fig:sub2}
    \end{subfigure}
    \hfill 
    \begin{subfigure}{0.24\textwidth}
        \includegraphics[width=\linewidth]{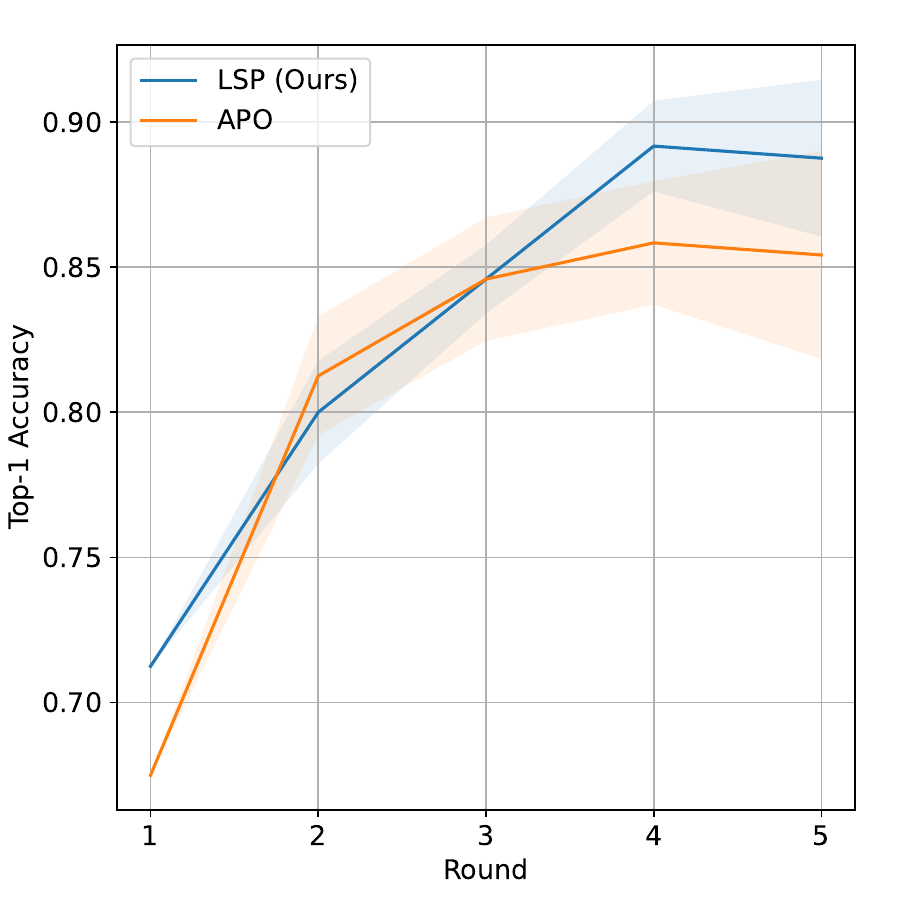}
        \caption{CUB-Blackbird}
        \label{fig:sub3}
    \end{subfigure}
    \hfill 
    \begin{subfigure}{0.24\textwidth}
        \includegraphics[width=\linewidth]{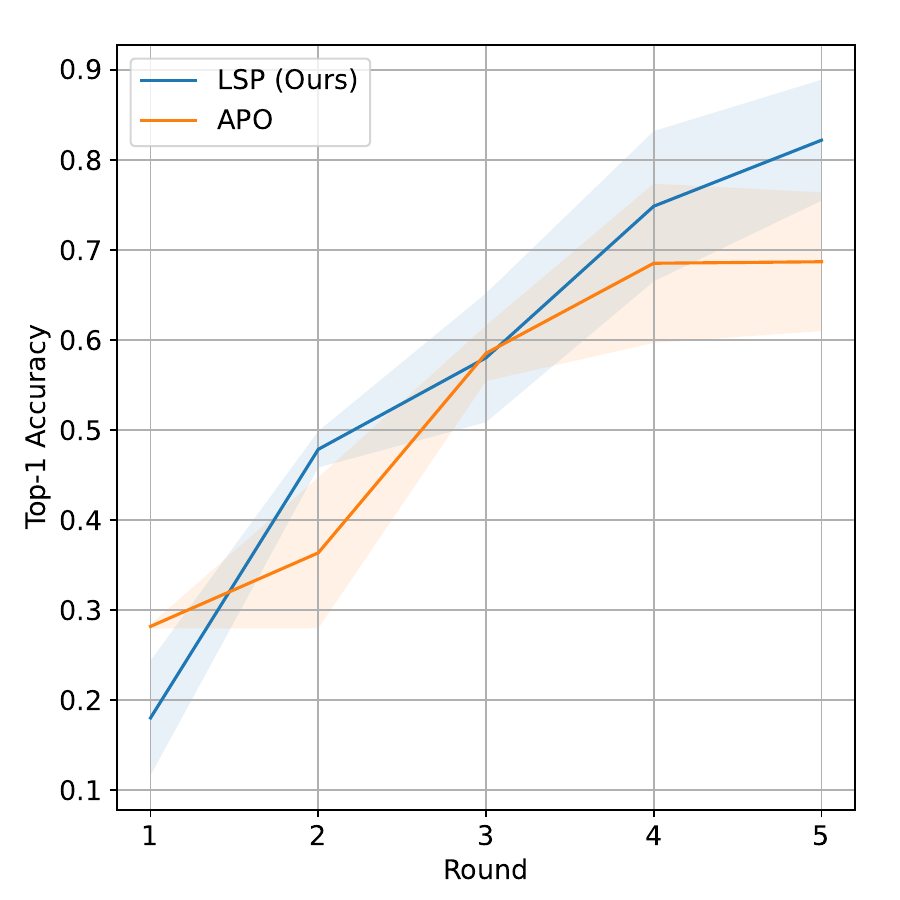}
        \caption{DT-Hard}
        \label{fig:sub4}
    \end{subfigure}
    \caption{
        \small
        \textbf{Convergence of different algorithms across time}. We plot the trajectory of training accuracy against the number of optimization rounds. The API model is GPT-3.5. \uline{(1). LSP converges substantially faster than vanilla prompting; (2). The search process does not introduce extra variances}.
        \vspace{-10pt}
    }
    \label{fig:learning_curves}
\end{figure}
\begin{wraptable}{r}{0.35\textwidth}
    \vspace{-4mm}
    \centering
    \caption{\small \textbf{Search and inference runtime Ccmparison.} Despite the multi-step decision-making process of LSP's tree structure, our method incurs comparable search and inference costs to various prompt optimization baselines.}
    \resizebox{0.3\textwidth}{!}{
        \begin{tabular}{lccccccc}
            
            \toprule
            \textbf{Method} & \textbf{Search (s)} & \textbf{Inference (s)} \\
            \midrule
            APE         & 270.60 & 0.11 \\
            OPRO        & 257.86 & 0.14 \\
            APO         & 270.85 & \textbf{0.08} \\
            PromptAgent & \textbf{220.95} & 0.11 \\
            LSP         & 232.54 & 0.13 \\
            \bottomrule
        \end{tabular}
    }
\vspace{-4mm}
\label{tab:ablate.cost}
\end{wraptable}
LSP organizes instructions into a tree-based structure.
Such divide-and-conquer strategy simplifies the learning process.
To verify this, we also plot the training trajectories for LSP across various tasks.
The training trajectory indicates the how fast a model fits the observed examples.
As Figure~\ref{fig:learning_curves} demonstrates, LSP not only converges faster but also achieves higher final accuracy compared to models that use unstructured prompting techniques.
\begin{wraptable}{r}{0.5\textwidth}
\vspace{-4mm}
    \centering
    \caption{\small \textbf{Comparison of Different Node Scoring Functions} on three tasks from IL-Bench-Language. Despite its simplicity, error count achieves more consistent performance compared to alternative metrics.}
    \resizebox{0.5\textwidth}{!}{
        \begin{tabular}{lccccccc}
        
            \toprule
            \textbf{Node Scoring} & \textbf{DT-Hard} & \textbf{Waxwing} & \textbf{Waterthrush} \\ \midrule \midrule
            Random          & 70.50 $\pm$ 11.01 & 62.22 $\pm$ 4.78 & 61.67 $\pm$ 1.36 \\
            Accuracy        & 80.33 $\pm$ 18.27 & \bf{66.11 $\pm$ 7.86} & 54.44 $\pm$ 0.70 \\
            Error Count (LSP)     & \bf{89.75 $\pm$ 1.25} & \bf{65.83 $\pm$ 4.17} & \bf{62.50 $\pm$ 0.83} \\
            \bottomrule
        \end{tabular}
    }
    
        

\vspace{-8mm}
\label{tab:ablate.components}
\end{wraptable}

\paragraph{Search cost analysis}
We also report the actual runtime of search and inference process for different methods in Table~\ref{tab:ablate.cost}.
We found that LSP incurs comparable search and inference costs to various prompt optimization baselines.

\paragraph{Different node scoring functions}
Table~\ref{tab:ablate.components} summarizes the performance of LSP using three different node scoring functions:
(1). Error count.
(2). Prediction accuracy.
(3). Random scoring.
The results suggest that Error count achieves substantially better and more consistent outcomes across different tasks.

\paragraph{Complexity of Learned LSP}
Our analysis of the statistics of learned programs indicates that the complexity of programs developed by LSP is quite manageable:
Most programs can reach a final prediction within just three steps, as illustrated in Figure~\ref{fig:tree_stats.depth}, and the tree structures tend to be sparse, as shown in  Figure~\ref{fig:tree_stats.sparsity}.
These observations confirm that although theoretical maximum tree expansion could grow exponentially with depth, in practice, LSPs operate effectively without requiring overly complex  structures.
This further explain why LSP exhibits the reasonable search and inference costs documented in Table~\ref{tab:ablate.cost}.

\vspace{-5pt}
\section{Conclusion}
\label{sec.conclusion}
\vspace{-5pt}
This work aims at revitalizing the concept of Neuro-Symbolic Programming in the era of Large Language Models.
We demonstrate that pretrained LLMs can implement powerful symbolic programs that are expressive, interpretable, and easy to train.
Additionally, we introduce the Instruction Learning Benchmark (IL-Benchmark), which consists of a suite of vision and language datasets designed to evaluate instruction learning algorithms.
We hope that our proposed framework will inspire new developments in interpretable learning methods during the LLM era.
We regard our study as an initial step in the research on LLM-Symbolic Programs.
Accordingly, we acknowledge the \textbf{limitations} of the current method in Appendix Section~\ref{sec:app.limitations}.

\bibliographystyle{plainnat}
\bibliography{main.bib}

\begin{thebibliography}{51}
\providecommand{\natexlab}[1]{#1}
\providecommand{\url}[1]{\texttt{#1}}
\expandafter\ifx\csname urlstyle\endcsname\relax
  \providecommand{\doi}[1]{doi: #1}\else
  \providecommand{\doi}{doi: \begingroup \urlstyle{rm}\Url}\fi

\bibitem[Achiam et~al.(2023)Achiam, Adler, Agarwal, Ahmad, Akkaya, Aleman, Almeida, Altenschmidt, Altman, Anadkat, et~al.]{gpt4}
Josh Achiam, Steven Adler, Sandhini Agarwal, Lama Ahmad, Ilge Akkaya, Florencia~Leoni Aleman, Diogo Almeida, Janko Altenschmidt, Sam Altman, Shyamal Anadkat, et~al.
\newblock Gpt-4 technical report.
\newblock \emph{arXiv preprint arXiv:2303.08774}, 2023.

\bibitem[Adams et~al.(2023)Adams, Fabbri, Ladhak, Lehman, and Elhadad]{summary1}
Griffin Adams, Alexander Fabbri, Faisal Ladhak, Eric Lehman, and No{\'e}mie Elhadad.
\newblock From sparse to dense: Gpt-4 summarization with chain of density prompting.
\newblock \emph{arXiv preprint arXiv:2309.04269}, 2023.

\bibitem[Ancona et~al.(2017)Ancona, Ceolini, {\"O}ztireli, and Gross]{ancona2017towards}
Marco Ancona, Enea Ceolini, Cengiz {\"O}ztireli, and Markus Gross.
\newblock Towards better understanding of gradient-based attribution methods for deep neural networks.
\newblock \emph{arXiv preprint arXiv:1711.06104}, 2017.

\bibitem[Bai et~al.(2023)Bai, Yeh, Ravikumar, Lin, and Hsieh]{bai2022concept}
Andrew Bai, Chih-Kuan Yeh, Pradeep Ravikumar, Neil~YC Lin, and Cho-Jui Hsieh.
\newblock Concept gradient: Concept-based interpretation without linear assumption.
\newblock In \emph{ICLR}, 2023.

\bibitem[Chaudhuri et~al.(2021)Chaudhuri, Ellis, Polozov, Singh, Solar-Lezama, Yue, et~al.]{nsp}
Swarat Chaudhuri, Kevin Ellis, Oleksandr Polozov, Rishabh Singh, Armando Solar-Lezama, Yisong Yue, et~al.
\newblock Neurosymbolic programming.
\newblock \emph{Foundations and Trends{\textregistered} in Programming Languages}, 7\penalty0 (3):\penalty0 158--243, 2021.

\bibitem[Chen and Guestrin(2016)]{xgboost}
Tianqi Chen and Carlos Guestrin.
\newblock Xgboost: A scalable tree boosting system.
\newblock In \emph{Proceedings of the 22nd acm sigkdd international conference on knowledge discovery and data mining}, pages 785--794, 2016.

\bibitem[Cui and Zhu(2021)]{dpad}
Guofeng Cui and He~Zhu.
\newblock Differentiable synthesis of program architectures.
\newblock \emph{Advances in Neural Information Processing Systems}, 34:\penalty0 11123--11135, 2021.

\bibitem[Dabkowski and Gal(2017)]{dabkowski2017real}
Piotr Dabkowski and Yarin Gal.
\newblock Real time image saliency for black box classifiers.
\newblock In \emph{Advances in Neural Information Processing Systems}, pages 6967--6976. NeurIPS, 2017.

\bibitem[Deng et~al.(2022)Deng, Wang, Hsieh, Wang, Guo, Shu, Song, Xing, and Hu]{rlprompt}
Mingkai Deng, Jianyu Wang, Cheng-Ping Hsieh, Yihan Wang, Han Guo, Tianmin Shu, Meng Song, Eric~P Xing, and Zhiting Hu.
\newblock Rlprompt: Optimizing discrete text prompts with reinforcement learning.
\newblock \emph{arXiv preprint arXiv:2205.12548}, 2022.

\bibitem[Dhurandhar et~al.(2018)Dhurandhar, Chen, Luss, Tu, Ting, Shanmugam, and Das]{dhurandhar2018explanations}
Amit Dhurandhar, Pin-Yu Chen, Ronny Luss, Chun-Chen Tu, Paishun Ting, Karthikeyan Shanmugam, and Payel Das.
\newblock Explanations based on the missing: Towards contrastive explanations with pertinent negatives.
\newblock In \emph{Advances in Neural Information Processing Systems}, pages 592--603. NeurIPS, 2018.

\bibitem[Fernando et~al.(2023)Fernando, Banarse, Michalewski, Osindero, and Rockt{\"a}schel]{promptbreeder}
Chrisantha Fernando, Dylan Banarse, Henryk Michalewski, Simon Osindero, and Tim Rockt{\"a}schel.
\newblock Promptbreeder: Self-referential self-improvement via prompt evolution.
\newblock \emph{arXiv preprint arXiv:2309.16797}, 2023.

\bibitem[Frosst and Hinton(2017)]{distill}
Nicholas Frosst and Geoffrey Hinton.
\newblock Distilling a neural network into a soft decision tree.
\newblock \emph{arXiv preprint arXiv:1711.09784}, 2017.

\bibitem[Goyal et~al.(2022)Goyal, Li, and Durrett]{summary2}
Tanya Goyal, Junyi~Jessy Li, and Greg Durrett.
\newblock News summarization and evaluation in the era of gpt-3.
\newblock \emph{arXiv preprint arXiv:2209.12356}, 2022.

\bibitem[Goyal et~al.(2019)Goyal, Wu, Ernst, Batra, Parikh, and Lee]{goyal2019counterfactual}
Yash Goyal, Ziyan Wu, Jan Ernst, Dhruv Batra, Devi Parikh, and Stefan Lee.
\newblock Counterfactual visual explanations.
\newblock In \emph{International Conference on Machine Learning}, pages 2376--2384. ICML, 2019.

\bibitem[Guo et~al.(2023)Guo, Wang, Guo, Li, Song, Tan, Liu, Bian, and Yang]{genetic}
Qingyan Guo, Rui Wang, Junliang Guo, Bei Li, Kaitao Song, Xu~Tan, Guoqing Liu, Jiang Bian, and Yujiu Yang.
\newblock Connecting large language models with evolutionary algorithms yields powerful prompt optimizers.
\newblock \emph{arXiv preprint arXiv:2309.08532}, 2023.

\bibitem[Hastie and Tibshirani(1990)]{gam}
Trevor Hastie and Robert Tibshirani.
\newblock \emph{Generalized additive models}.
\newblock Chapman and Hall/CRC, 1990.

\bibitem[Hendricks et~al.(2018)Hendricks, Hu, Darrell, and Akata]{Hendricks2018GroundingVE}
Lisa~Anne Hendricks, Ronghang Hu, Trevor Darrell, and Zeynep Akata.
\newblock Grounding visual explanations.
\newblock In \emph{ECCV}. ECCV, 2018.

\bibitem[Hsieh et~al.(2021)Hsieh, Yeh, Liu, Ravikumar, Kim, Kumar, and Hsieh]{hsieh2021evaluations}
Cheng-Yu Hsieh, Chih-Kuan Yeh, Xuanqing Liu, Pradeep~Kumar Ravikumar, Seungyeon Kim, Sanjiv Kumar, and Cho-Jui Hsieh.
\newblock Evaluations and methods for explanation through robustness analysis.
\newblock In \emph{International Conference on Learning Representations}. ICLR, 2021.
\newblock URL \url{https://openreview.net/forum?id=4dXmpCDGNp7}.

\bibitem[Hsieh et~al.(2023)Hsieh, Si, Yu, and Dhillon]{long_prompt}
Cho-Jui Hsieh, Si~Si, Felix~X Yu, and Inderjit~S Dhillon.
\newblock Automatic engineering of long prompts.
\newblock \emph{arXiv preprint arXiv:2311.10117}, 2023.

\bibitem[Kim et~al.(2018)Kim, Wattenberg, Gilmer, Cai, Wexler, Viegas, et~al.]{kim2018interpretability}
Been Kim, Martin Wattenberg, Justin Gilmer, Carrie Cai, James Wexler, Fernanda Viegas, et~al.
\newblock Interpretability beyond feature attribution: Quantitative testing with concept activation vectors (tcav).
\newblock In \emph{International Conference on Machine Learning}, pages 2673--2682. ICML, 2018.

\bibitem[Koh et~al.(2020)Koh, Nguyen, Tang, Mussmann, Pierson, Kim, and Liang]{koh2020concept}
Pang~Wei Koh, Thao Nguyen, Yew~Siang Tang, Stephen Mussmann, Emma Pierson, Been Kim, and Percy Liang.
\newblock Concept bottleneck models.
\newblock In \emph{International conference on machine learning}, pages 5338--5348. PMLR, 2020.

\bibitem[Kramberger and Poto{\v{c}}nik(2020)]{cars}
Tin Kramberger and Bo{\v{z}}idar Poto{\v{c}}nik.
\newblock Lsun-stanford car dataset: enhancing large-scale car image datasets using deep learning for usage in gan training.
\newblock \emph{Applied Sciences}, 10\penalty0 (14):\penalty0 4913, 2020.

\bibitem[Losch et~al.(2019)Losch, Fritz, and Schiele]{losch2019interpretability}
Max Losch, Mario Fritz, and Bernt Schiele.
\newblock Interpretability beyond classification output: Semantic bottleneck networks.
\newblock \emph{arXiv preprint arXiv:1907.10882}, 2019.

\bibitem[Maji et~al.(2013)Maji, Rahtu, Kannala, Blaschko, and Vedaldi]{fgvc}
Subhransu Maji, Esa Rahtu, Juho Kannala, Matthew Blaschko, and Andrea Vedaldi.
\newblock Fine-grained visual classification of aircraft.
\newblock \emph{arXiv preprint arXiv:1306.5151}, 2013.

\bibitem[Ming et~al.(2019)Ming, Xu, Qu, and Ren]{prototype1}
Yao Ming, Panpan Xu, Huamin Qu, and Liu Ren.
\newblock Interpretable and steerable sequence learning via prototypes.
\newblock In \emph{Proceedings of the 25th ACM SIGKDD International Conference on Knowledge Discovery \& Data Mining}, pages 903--913, 2019.

\bibitem[Nauta et~al.(2021{\natexlab{a}})Nauta, Jutte, Provoost, and Seifert]{prototype2}
Meike Nauta, Annemarie Jutte, Jesper Provoost, and Christin Seifert.
\newblock This looks like that, because... explaining prototypes for interpretable image recognition.
\newblock In \emph{Joint European Conference on Machine Learning and Knowledge Discovery in Databases}, pages 441--456. Springer, 2021{\natexlab{a}}.

\bibitem[Nauta et~al.(2021{\natexlab{b}})Nauta, Van~Bree, and Seifert]{prototree}
Meike Nauta, Ron Van~Bree, and Christin Seifert.
\newblock Neural prototype trees for interpretable fine-grained image recognition.
\newblock In \emph{Proceedings of the IEEE/CVF conference on computer vision and pattern recognition}, pages 14933--14943, 2021{\natexlab{b}}.

\bibitem[Nilsback and Zisserman(2008)]{flowers}
Maria-Elena Nilsback and Andrew Zisserman.
\newblock Automated flower classification over a large number of classes.
\newblock In \emph{2008 Sixth Indian conference on computer vision, graphics \& image processing}, pages 722--729. IEEE, 2008.

\bibitem[Oikarinen et~al.(2023)Oikarinen, Das, Nguyen, and Weng]{oikarinen2023label}
Tuomas Oikarinen, Subhro Das, Lam~M Nguyen, and Tsui-Wei Weng.
\newblock Label-free concept bottleneck models.
\newblock \emph{arXiv preprint arXiv:2304.06129}, 2023.

\bibitem[Ouyang et~al.(2022)Ouyang, Wu, Jiang, Almeida, Wainwright, Mishkin, Zhang, Agarwal, Slama, Ray, et~al.]{gpt3}
Long Ouyang, Jeffrey Wu, Xu~Jiang, Diogo Almeida, Carroll Wainwright, Pamela Mishkin, Chong Zhang, Sandhini Agarwal, Katarina Slama, Alex Ray, et~al.
\newblock Training language models to follow instructions with human feedback.
\newblock \emph{Advances in neural information processing systems}, 35:\penalty0 27730--27744, 2022.

\bibitem[Pair(2024)]{palworld}
Pocket Pair.
\newblock Palworld, 2024.
\newblock URL \url{https://en.wikipedia.org/wiki/Palworld}.

\bibitem[Petsiuk et~al.(2018)Petsiuk, Das, and Saenko]{petsiuk2018rise}
Vitali Petsiuk, Abir Das, and Kate Saenko.
\newblock Rise: Randomized input sampling for explanation of black-box models.
\newblock \emph{arXiv preprint arXiv:1806.07421}, 2018.

\bibitem[Pryzant et~al.(2023)Pryzant, Iter, Li, Lee, Zhu, and Zeng]{apo}
Reid Pryzant, Dan Iter, Jerry Li, Yin~Tat Lee, Chenguang Zhu, and Michael Zeng.
\newblock Automatic prompt optimization with" gradient descent" and beam search.
\newblock \emph{arXiv preprint arXiv:2305.03495}, 2023.

\bibitem[Pu and Demberg(2023)]{summary4}
Dongqi Pu and Vera Demberg.
\newblock Chatgpt vs human-authored text: Insights into controllable text summarization and sentence style transfer.
\newblock \emph{arXiv preprint arXiv:2306.07799}, 2023.

\bibitem[Ribeiro et~al.(2016)Ribeiro, Singh, and Guestrin]{lime}
Marco~Tulio Ribeiro, Sameer Singh, and Carlos Guestrin.
\newblock Why should i trust you?: Explaining the predictions of any classifier.
\newblock In \emph{Proceedings of the 22nd ACM SIGKDD international conference on knowledge discovery and data mining}, pages 1135--1144. ACM, 2016.

\bibitem[Shah et~al.(2020)Shah, Zhan, Sun, Verma, Yue, and Chaudhuri]{near}
Ameesh Shah, Eric Zhan, Jennifer Sun, Abhinav Verma, Yisong Yue, and Swarat Chaudhuri.
\newblock Learning differentiable programs with admissible neural heuristics.
\newblock \emph{Advances in neural information processing systems}, 33:\penalty0 4940--4952, 2020.

\bibitem[Shin et~al.(2020)Shin, Razeghi, Logan~IV, Wallace, and Singh]{autoprompt}
Taylor Shin, Yasaman Razeghi, Robert~L Logan~IV, Eric Wallace, and Sameer Singh.
\newblock Autoprompt: Eliciting knowledge from language models with automatically generated prompts.
\newblock \emph{arXiv preprint arXiv:2010.15980}, 2020.

\bibitem[Shrikumar et~al.(2017)Shrikumar, Greenside, and Kundaje]{shrikumar2017learning}
Avanti Shrikumar, Peyton Greenside, and Anshul Kundaje.
\newblock Learning important features through propagating activation differences.
\newblock \emph{International Conference on Machine Learning}, 2017.

\bibitem[Sundararajan et~al.(2017)Sundararajan, Taly, and Yan]{sundararajan2017axiomatic}
Mukund Sundararajan, Ankur Taly, and Qiqi Yan.
\newblock Axiomatic attribution for deep networks.
\newblock In \emph{International Conference on Machine Learning}, pages 3319--3328. PMLR, 2017.

\bibitem[Team et~al.(2023)Team, Anil, Borgeaud, Wu, Alayrac, Yu, Soricut, Schalkwyk, Dai, Hauth, et~al.]{gemini}
Gemini Team, Rohan Anil, Sebastian Borgeaud, Yonghui Wu, Jean-Baptiste Alayrac, Jiahui Yu, Radu Soricut, Johan Schalkwyk, Andrew~M Dai, Anja Hauth, et~al.
\newblock Gemini: a family of highly capable multimodal models.
\newblock \emph{arXiv preprint arXiv:2312.11805}, 2023.

\bibitem[van~der Waa et~al.(2018)van~der Waa, Robeer, van Diggelen, Brinkhuis, and Neerincx]{vanderwaa2018}
Jasper van~der Waa, Marcel Robeer, Jurriaan van Diggelen, Matthieu Brinkhuis, and Mark Neerincx.
\newblock {Contrastive Explanations with Local Foil Trees}.
\newblock In \emph{2018 Workshop on Human Interpretability in Machine Learning (WHI)}. WHI, 2018.

\bibitem[Van~Horn et~al.(2015)Van~Horn, Branson, Farrell, Haber, Barry, Ipeirotis, Perona, and Belongie]{nabirds}
Grant Van~Horn, Steve Branson, Ryan Farrell, Scott Haber, Jessie Barry, Panos Ipeirotis, Pietro Perona, and Serge Belongie.
\newblock Building a bird recognition app and large scale dataset with citizen scientists: The fine print in fine-grained dataset collection.
\newblock In \emph{Proceedings of the IEEE conference on computer vision and pattern recognition}, pages 595--604, 2015.

\bibitem[Wah et~al.(2011)Wah, Branson, Welinder, Perona, and Belongie]{cub}
C.~Wah, S.~Branson, P.~Welinder, P.~Perona, and S.~Belongie.
\newblock The caltech-ucsd birds-200-2011 dataset.
\newblock Technical Report CNS-TR-2011-001, California Institute of Technology, 2011.

\bibitem[Wang et~al.(2023)Wang, Li, Wang, Bai, Luo, Zhang, Jojic, Xing, and Hu]{promptagent}
Xinyuan Wang, Chenxi Li, Zhen Wang, Fan Bai, Haotian Luo, Jiayou Zhang, Nebojsa Jojic, Eric~P Xing, and Zhiting Hu.
\newblock Promptagent: Strategic planning with language models enables expert-level prompt optimization.
\newblock \emph{arXiv preprint arXiv:2310.16427}, 2023.

\bibitem[Xu et~al.(2022)Xu, Chen, Du, Shao, Wang, Li, and Yang]{gps}
Hanwei Xu, Yujun Chen, Yulun Du, Nan Shao, Yanggang Wang, Haiyu Li, and Zhilin Yang.
\newblock Gps: Genetic prompt search for efficient few-shot learning.
\newblock \emph{arXiv preprint arXiv:2210.17041}, 2022.

\bibitem[Yang et~al.(2023)Yang, Wang, Lu, Liu, Le, Zhou, and Chen]{opro}
Chengrun Yang, Xuezhi Wang, Yifeng Lu, Hanxiao Liu, Quoc~V Le, Denny Zhou, and Xinyun Chen.
\newblock Large language models as optimizers.
\newblock \emph{arXiv preprint arXiv:2309.03409}, 2023.

\bibitem[Yuksekgonul et~al.(2022)Yuksekgonul, Wang, and Zou]{yuksekgonul2022post}
Mert Yuksekgonul, Maggie Wang, and James Zou.
\newblock Post-hoc concept bottleneck models.
\newblock \emph{arXiv preprint arXiv:2205.15480}, 2022.

\bibitem[Zhang et~al.(2022)Zhang, Wang, Zhou, Schuurmans, and Gonzalez]{tempera}
Tianjun Zhang, Xuezhi Wang, Denny Zhou, Dale Schuurmans, and Joseph~E Gonzalez.
\newblock Tempera: Test-time prompting via reinforcement learning.
\newblock \emph{arXiv preprint arXiv:2211.11890}, 2022.

\bibitem[Zhang et~al.(2024)Zhang, Ladhak, Durmus, Liang, McKeown, and Hashimoto]{summary3}
Tianyi Zhang, Faisal Ladhak, Esin Durmus, Percy Liang, Kathleen McKeown, and Tatsunori~B Hashimoto.
\newblock Benchmarking large language models for news summarization.
\newblock \emph{Transactions of the Association for Computational Linguistics}, 12:\penalty0 39--57, 2024.

\bibitem[Zhou et~al.(2022)Zhou, Muresanu, Han, Paster, Pitis, Chan, and Ba]{ape}
Yongchao Zhou, Andrei~Ioan Muresanu, Ziwen Han, Keiran Paster, Silviu Pitis, Harris Chan, and Jimmy Ba.
\newblock Large language models are human-level prompt engineers.
\newblock \emph{arXiv preprint arXiv:2211.01910}, 2022.

\bibitem[Zintgraf et~al.(2017)Zintgraf, Cohen, Adel, and Welling]{zintgraf2017visualizing}
Luisa~M Zintgraf, Taco~S Cohen, Tameem Adel, and Max Welling.
\newblock Visualizing deep neural network decisions: Prediction difference analysis.
\newblock \emph{arXiv preprint arXiv:1702.04595}, 2017.

\end{thebibliography}


\newpage
\section{Supplemental Material}
\label{sec.supp}

\subsection{Additional ablation experiments}

\subsubsection{Using different LLMs to implement LSPs}
\label{sec:exp.cross_model}
The role of LLMs in LSPs is twofold: they serve both as the inference and learning engine of the LLM-modules in the grammar.
The learning engine is responsible for summarizing and organizing patterns from observed data samples into clear predictive rules, whereas the inference engine follows the learned program to make predictions on test examples.
Natural questions arise: (1). how effective are different LLMs at optimizing LSPs? (2). Is the learned programs interpretable to different LLMs?

\paragraph{LLM as LSP learner}
We replace the learning engine used in optimizing LSP with various LLMs - GPT-3.5, Gemini, and GPT-4 - while keeping all other settings consistent with the main experiment.
As shown in Figure~\ref{fig:optimizer}, GPT-4 consistently outperforms other LLMs on both text and vision tasks, while Gemini and GPT-3.5 show similar performance with each other.
This reflects their respective capabilities.
For specific examples of instructions generated by different LLM optimizers, please see the Appendix.

\paragraph{LLM as LSP interpreter}
\begin{table}[ht]

    \caption{\small \textbf{Transferring LSPs learned from one LLM to another.} The learned LSPs are generally interpretable across various LLMs. However, larger LLMs (e.g., GPT-4) demonstrate a slightly higher consistency in understanding LSPs learned by other LLMs.}

    \centering
    \resizebox{0.6\textwidth}{!}{
    \begin{tabular}{cccccc}
        \toprule
        \multirow{2}{*}{Source Model} & \multirow{2}{*}{Task} & \multicolumn{4}{c}{Evaluator} \\
        \cmidrule{3-6}
        & & GPT3.5 & Gemini-M & GPT4 & \\
        \midrule
        
        \multirow{4}{*}{GPT3.5}
            & DT-Hard & 89.75 $\pm$ 1.25 & 72.67 $\pm$ 6.91 & 87.50 $\pm$ 1.22 \\
            \cmidrule(r){2-6}
            & Waxwing & 65.83 $\pm$ 4.17 & 52.22 $\pm$ 1.57 & 56.67 $\pm$ 3.60 \\
            \cmidrule(r){2-6}
            & Waterthrush & 62.50 $\pm$ 0.83 & 64.44 $\pm$ 0.79 & 59.44 $\pm$ 3.93 \\
        \midrule
        
        \multirow{4}{*}{Gemini-M}
            & DT-Hard & 75.50 $\pm$ 2.04 & 80.83 $\pm$ 1.03 & 79.17 $\pm$ 11.45 \\
            \cmidrule(r){2-6}
            & Waxwing & 52.78 $\pm$ 3.42 & 58.33 $\pm$ 4.91 & 61.11 $\pm$ 10.57 \\
            \cmidrule(r){2-6}
            & Waterthrush & 50.56 $\pm$ 4.16 & 54.44 $\pm$ 5.50 & 52.22 $\pm$ 0.79 \\
        \midrule
        
        \multirow{4}{*}{GPT4}
            & DT-Hard & 74.50 $\pm$ 9.35 & 57.67 $\pm$ 3.01 & 99.50 $\pm$ 0.00 \\
            \cmidrule(r){2-6}
            & Waxwing & 59.44 $\pm$ 5.15 & 62.22 $\pm$ 7.49 & 63.33 $\pm$ 4.91 \\
            \cmidrule(r){2-6}
            & Waterthrush & 66.67 $\pm$ 6.80 & 68.33 $\pm$ 2.72 & 62.78 $\pm$ 9.06 \\
        \bottomrule
    
    \end{tabular}
    }
\vspace{-2mm}
\label{tab:ablate.transfer}
\end{table}

We then test if LSPs created by one LLM could be interpreted by other LLMs.
Table~\ref{tab:ablate.transfer} summarizes the performance.
The results suggest that LSPs are interpretable across a diverse range of inference models;
Larger and stronger LLMs (e.g. GPT-4) demonstrates a slight more consistent ability in interpreting LSPs, which aligns their superior instruction-following capacities.

\subsection{Learning algorithm for LSP}
The complete pipeline for constructing LSP is summarized in Algorithm~\ref{algo:learn_llm_module} and Algorithm~\ref{algo:search_algorithm}.

\paragraph{Remarks}
\begin{itemize}[noitemsep, topsep=0pt, parsep=0pt, partopsep=0pt, leftmargin=*]
    \item Although initially, the complexity of the program expansion might seem exponential to the tree depth, a closer examination reveals otherwise: (1). In practice, the trees are typically sparse, meaning that expanding only a few branches is often sufficient to achieve good performance (Figure~\ref{fig:tree_stats.sparsity}). (2). The divide-and-conquer approach ensures that each tree level processes the same amount of data making the evaluation complexity linear to tree depth.
    \item The above arrangement of the search process does not compromise generality of LSP: For more sophisticated DSL designs, program structure search can be conducted similarly to traditional NSPs, using top-down tree traversal (cite).
\end{itemize}

\begin{algorithm}
\caption{\texttt{learn\_llm\_module}: Learning LLM Module by summarizing predictive rules}
\label{algo:learn_llm_module}
    \begin{algorithmic}[1]
    \STATE {\bfseries Input:} Proposal size $m$, data sample $\mathcal{B}$, learner LLM $\mathcal{M}_l$
    \STATE Initialize an empty list of LLM modules $\Phi$
    \FOR{$i = 1$ \TO $m$}
        \STATE Randomly sample $b \sim \mathcal{B}$
        \STATE $\phi_{new} \leftarrow \texttt{summarize}(M_l, b)$
        \STATE $\Phi \leftarrow \Phi \cup \{\phi_{new}\}$
    \ENDFOR
    \RETURN $\Phi$
    \end{algorithmic}
\end{algorithm}

\begin{algorithm}
\caption{Complete pipeline of optimizing LSPs}
\label{algo:search_algorithm}
    \begin{algorithmic}[1]
    \STATE {\bfseries Input:} Dataset $\mathcal{D}$, beam size $d$, number of iterations $T$, inference LLM $\mathcal{M}_i$, learner LLM $\mathcal{M}_l$, expand ratio $K$, proposal size $m$
    
    \vspace{0.3em}
    \STATE Initialize $p_0$ as an empty program
    \STATE Initialize candidate program set $P = \{p_0\}$
    
    \vspace{0.3em}
    \FOR{$t = 1$ \TO $T$}
        \FOR{each program $p$ in $P$}
            \vspace{0.3em}
            \STATE \textit{\color{cyan} $\triangleright$ Batch evaluation}
            \STATE Sample a batch $\mathcal{B} \sim \mathcal{D}$
            \STATE Evaluate $p$ on $\mathcal{B}$ using $\mathcal{M}_i$
            
            \vspace{0.3em}
            \STATE \textit{\color{cyan} $\triangleright$ Selecting the most promising node $n$ to expand}
            \STATE Assign $\mathcal{B}$ to the leaf nodes of $p$
            \STATE Identify the most error-prone leaf node $n$ with assigned subset $\mathcal{B}_n$
            
            \vspace{0.3em}
            \STATE \textit{\color{cyan} $\triangleright$ Extend program $p$ to $K$ new programs by adding top-$K$ LLM modules to node $n$}
            \STATE $\Phi \leftarrow \texttt{learn\_llm\_module}(n, \mathcal{B}_n, \mathcal{M}_l, m)$
            \STATE $\Phi_{topK} \leftarrow$ evaluate and retain top-$K$ $\Phi$ on $\mathcal{B}_n$
            \STATE $\mathcal{P}_{new} \leftarrow$ extend $p$ by assigning each $\phi \in \Phi_{topK}$ to node $n$ on program $p$.
            \STATE $\mathcal{P} \leftarrow \mathcal{P} \cup \mathcal{P}_{new}$
        \ENDFOR
        \STATE Evaluate and retain the top-$d$ programs from $\mathcal{P}$ on $\mathcal{D}$
    \ENDFOR
    \RETURN The best program from $P$
    \end{algorithmic}
\end{algorithm}

\begin{table}[t!]
\tiny
\centering
\caption{\textbf{Overview of Interpretable-Learning Benchmark}. We provide task names, types, summaries, number of labels, and one example data point for each task.}

\resizebox{0.99\textwidth}{!}{
    \begin{tabular}{p{1.1cm}p{1cm}p{3cm}p{0.5cm}p{8cm}}

    \toprule
    \textbf{Task} & \textbf{Type} & \textbf{Summary} & \textbf{Labels} & \textbf{Example} \\
    \midrule
    \midrule
    
    DT-Easy & Synthetic
    & Predict labels based on symbolic inputs. Rules generated by a small decision tree
    & 2
    & "input": "x1=A2; x2=B1", "output": "bar" \\
    \midrule

    DT-Medium & Synthetic
    & Predict labels based on symbolic inputs. Rules generated by a medium decision tree
    & 2
    & "input": "x1=A3; x2=B2", "output": "bar" \\
    \midrule

    DT-Hard & Synthetic
    & Predict labels based on symbolic inputs. Rules generated by a large decision tree
    & 4
    & "input": "x1=A1; x2=B1; x3=C1", "output": "foo" \\
    \midrule
    
    \midrule
    
    Waxwing & Caption
    & Classify Waxwing species based on its text description.
    & 2
    & "input": "Tan to light brown head and upper body, black \"mask\" across eyes, lighter cream underparts, bright red tips on secondary wing feathers, small black bill, yellow band on tail.", "output": "Cedar Waxwing" \\
    \midrule

    Waterthrush & Caption
    & Classify Waterthrush species based on its text description.
    & 2
    & "input": "Light gray crown, white supercilium, dark eyestripe extending behind eye, olive-brown wings with faint wingbars, white throat, pale underparts, long, slender bill, relatively short tail, orange legs.", "output": "Louisiana Waterthrush" \\
    \midrule

    Jaeger & Caption
    & Classify Jaeger species based on its text description.
    & 2
    & "input": "Light greyish-brown plumage on the underside, distinct narrow white band across the nape, wings with a M-shaped pattern when spread, tail slightly forked but mostly straight across.", "output": "Long tailed Jaeger" \\
    \midrule

    Albatross & Caption
    & Classify Albatross species based on its text description.
    & 3
    & "input": "Dark brown upperparts and paler brown underparts, elongated and narrow wings with a white trailing edge and distinct finger-like tips, hooked beak with a pale base, light-colored head with a dark eye patch and bill, wings held straight in gliding flight, gliding above water surface. Uniform dark brown plumage, long slender wings, distinct white pattern on underwings, white band near the tips of the underwings, pale or white head, dark eye patch.", "output": "Black footed Albatross" \\
    \midrule

    Blackbird & Caption
    & Classify Blackbird species based on its text description.
    & 4
    & "input": "Bright yellow head, black body, sharp conical beak, perched on reed-like vegetation. Bright yellow head, yellow chest, solid black body excluding head and chest, perched on a thin branch. Black body, bright yellow head, sturdy bill, perched on a reed.", "output": "Yellow headed Blackbird" \\
    \midrule

    Swallow & Caption
    & Classify Swallow species based on its text description.
    & 4
    & "input": "Light brown head, pale throat, light brown upperparts, long pointed wings, short tail, white underparts, sitting on wire. Light brown head and upper body, white underparts, sitting on a wire, sky background, short beak, sleek body shape. Brown and white plumage, perched on a wire, stout body, short and thick neck, medium-length tail with a straight edge, compact size, unmarked lighter underparts, darker wings and upperparts.", "output": "Bank Swallow" \\
    \midrule
    
    \midrule
    
    Fire-1 & Vision
    & Distinguish visually-similar fire-type pals from Palworld.
    & 3
    & "input": \quad\quad\quad \makebox[0pt]{\raisebox{-0.5\height}{\includegraphics[width=0.75cm]{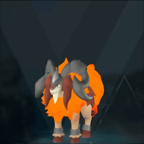}}}, \quad\quad\quad "output:" "Arsox" \\
    \midrule
        
    Fire-2 & Vision
    & Distinguish visually-similar fire-type pals from Palworld.
    & 5
    & "input": \quad\quad\quad \makebox[0pt]{\raisebox{-0.5\height}{\includegraphics[width=0.75cm]{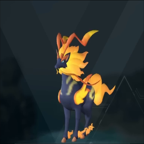}}}, \quad\quad\quad "output:" "Pyrin" \\
    \midrule
    
    Dragon-Blue-1 & Vision
    & Distinguish visually-similar blue-colored dragon-type pals from Palworld.
    & 3
    & "input": \quad\quad\quad \makebox[0pt]{\raisebox{-0.5\height}{\includegraphics[width=0.75cm]{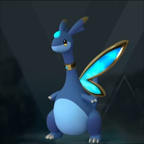}}}, \quad\quad\quad "output:" "Elphidran Aqua" \\
    \midrule
    
    Dragon-Blue-2 & Vision
    & Distinguish visually-similar blue-colored dragon-type pals from Palworld.
    & 4
    & "input": \quad\quad\quad \makebox[0pt]{\raisebox{-0.5\height}{\includegraphics[width=0.75cm]{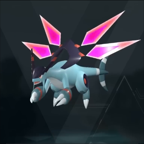}}}, \quad\quad\quad "output:" "Jetragon" \\
    \midrule
    
    Electric-1 & Vision
    & Distinguish visually-similar electric-type pals from Palworld.
    & 3
    & "input": \quad\quad\quad \makebox[0pt]{\raisebox{-0.5\height}{\includegraphics[width=0.75cm]{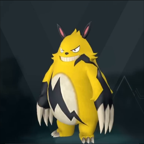}}}, \quad\quad\quad "output:" "Grizzbolt" \\
    \midrule
    
    Electric-2 & Vision
    & Distinguish visually-similar electric-type pals from Palworld.
    & 4
    & "input": \quad\quad\quad \makebox[0pt]{\raisebox{-0.5\height}{\includegraphics[width=0.75cm]{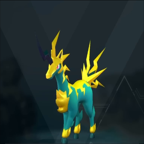}}}, \quad\quad\quad "output:" "Univolt" \\
    \midrule
    
    Water-1 & Vision
    & Distinguish visually-similar water-type pals from Palworld.
    & 4
    & "input": \quad\quad\quad \makebox[0pt]{\raisebox{-0.5\height}{\includegraphics[width=0.75cm]{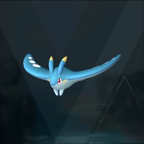}}}, \quad\quad\quad "output:" "Celaray" \\
    \midrule

    \bottomrule
    \end{tabular}
}
\label{tab:il_bench_overview}
\end{table}


\subsection{More details on empirical evaluation}
Here we provide more details on the empirical evaluation.
\subsubsection{Implementation details}
Throughout our main experiment, we use an expansion ratio of 4, batch size of 64, a maximum number of four iterations, and a maximum of 8 candidate (LLM module) proposals for each iteration.
The settings for beam search follows that of APO, which uses a beam size of 4 and deploys UCBBandits algorithm with a sample size of 32 to speedup the candidate ranking~\cite{apo}.
The only exception is that for vision tasks, we use a batch size of 4 for cost reduction.
The temperature for all API models are set to their default (0.7).

For all prompt optimization baselines, we set the maximum budget (measured by the number of candidate proposals) to the same number.
For Decision Tree, we use XGBoost library's standard implementation, which operates on raw pixels.
For ProtoTree, we directly run the original implementation, but reduce the maximum depth from 9 to 5, as it is faster to train yet achieves better performance on our datasets.
We align the evaluation our baselines

\subsubsection{Constructing Out-Of-Distribution dataset for IL-Bench-Vision tasks}

\begin{figure}[H]
    \centering
    \begin{subfigure}[b]{0.24\textwidth}
        \centering
        \includegraphics[width=\textwidth]{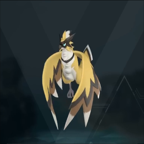}
        \caption{Beakon Original}
    \end{subfigure}
    \hfill
    \begin{subfigure}[b]{0.24\textwidth}
        \centering
        \includegraphics[width=\textwidth]{Figures/appendix/palworld/Celaray_frame_0445.png}
        \caption{Celaray Original}
    \end{subfigure}
    \hfill
    \begin{subfigure}[b]{0.24\textwidth}
        \centering
        \includegraphics[width=\textwidth]{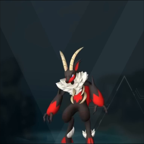}
        \caption{Incineram Original}
    \end{subfigure}
    \hfill
    \begin{subfigure}[b]{0.24\textwidth}
        \centering
        \includegraphics[width=\textwidth]{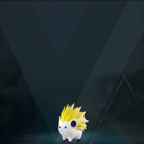}
        \caption{Jolthog Original}
    \end{subfigure}

    \vskip\baselineskip 

    \begin{subfigure}[b]{0.24\textwidth}
        \centering
        \includegraphics[width=\textwidth]{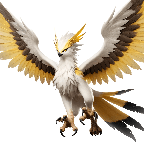}
        \caption{Beakon Generated}
    \end{subfigure}
    \hfill
    \begin{subfigure}[b]{0.24\textwidth}
        \centering
        \includegraphics[width=\textwidth]{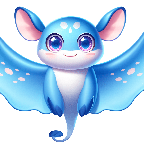}
        \caption{Celaray Generated}
    \end{subfigure}
    \hfill
    \begin{subfigure}[b]{0.24\textwidth}
        \centering
        \includegraphics[width=\textwidth]{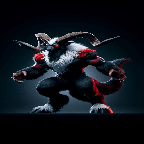}
        \caption{Incineram Generated}
    \end{subfigure}
    \hfill
    \begin{subfigure}[b]{0.24\textwidth}
        \centering
        \includegraphics[width=\textwidth]{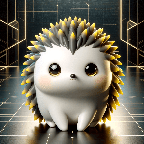}
        \caption{Jolthog Generated}
    \end{subfigure}

    \caption{\textbf{Comparison between original images (top row) and Out-Of-Distribution images (botton row) generated by GPT-4V. } All images are resized to an unified resolution of 128.}
    \label{fig:ood_example}
\end{figure}
Our OOD dataset is constructed by feeding the original image from the training set to GPT-4 (web version), and ask GPT to generate a variant of the input image.
The prompt we used is shown below.
Figure~\ref{fig:ood_example} shows a comparison of some example OOD images generated by GPT-4 with original image.

\begin{center}
    \begin{quote}
        \texttt{Generate an image variant containing the creature in the provided image. keep the key features of this creature unmodified. You must show the full body view of this creature.}
    \end{quote}
\end{center}

\subsubsection{Human Evaluation Protocol}
We conduct user study to access the interpretability of our method and ProtoTree.
For both methods, we send (1) the original image datasets and (2) visualizations of the discovered programs to the human raters, and as the human rater to make predictions based on those programs.
We then compute the accuracy of their predictions, and report the mean and standard deviations.
We select the group of human raters so that they have no background in machine learning research.
    
    

\begin{figure}[htbp]
    \centering
    
    \begin{subfigure}[b]{0.2\textwidth}
        \centering
        \includegraphics[width=\textwidth]{Figures/appendix/palworld/Celaray_frame_0445.png}
        \caption{Celaray}
        \label{fig:side1}
    \end{subfigure}
    \hfill
    \begin{subfigure}[b]{0.2\textwidth}
        \centering
        \includegraphics[width=\textwidth]{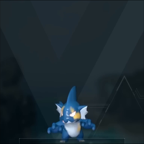}
        \caption{Gobfin}
        \label{fig:side2}
    \end{subfigure}
    \hfill
    \begin{subfigure}[b]{0.2\textwidth}
        \centering
        \includegraphics[width=\textwidth]{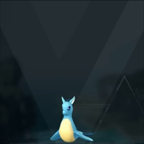}
        \caption{Kelpsea}
        \label{fig:side3}
    \end{subfigure}
    \hfill
    \begin{subfigure}[b]{0.2\textwidth}
        \centering
        \includegraphics[width=\textwidth]{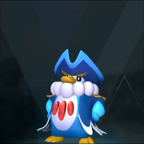}
        \caption{Penking}
        \label{fig:side4}
    \end{subfigure}
    \vfill
    
    \begin{subfigure}[b]{\textwidth}
        \centering
        \includegraphics[width=1.0\textwidth]{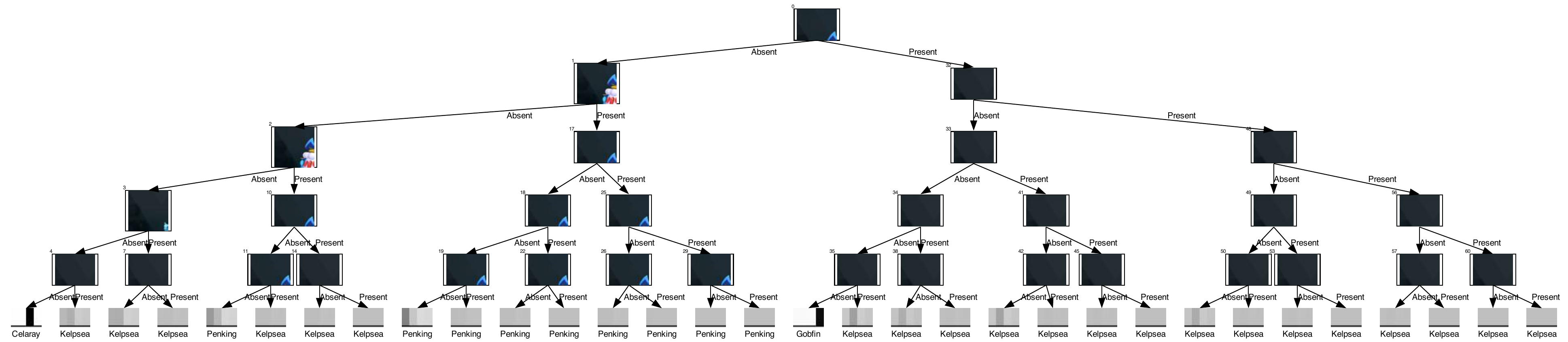}
        \caption{Caption for the second image}
        \label{fig:bottom}
    \end{subfigure}
    \vfill
    
    \begin{subfigure}[b]{\textwidth}
    \centering
    \includegraphics[width=1.0\textwidth]{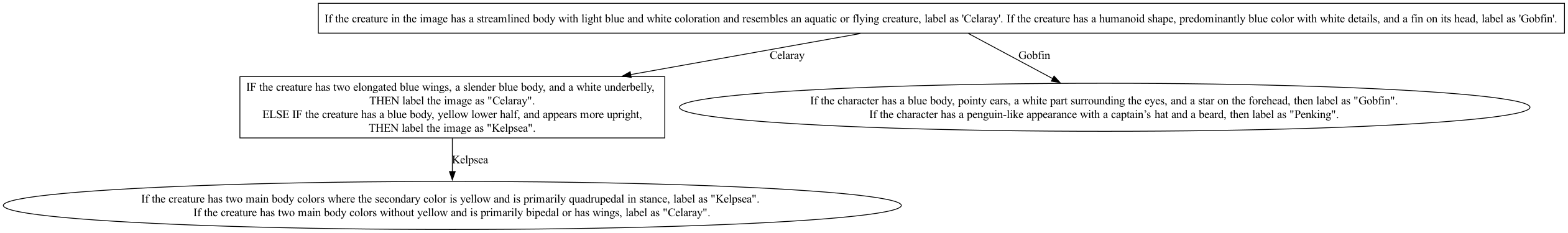}
    \caption{Example}
    \label{fig:top}
    \end{subfigure}
    \vfill
    
    \caption{\textbf{Example programs discovered by LSP (bottom) and ProtoTree (middle).} While ProtoTree offers some interpretability by displaying prototype image patches to the user, it can be misleading as there is no guarantee that the prototypes are meaningful (e.g. many patches miss the key regions, and there also exists entire branches that overfit to the background). In contrast, the programs discovered by LSP accurately capture the characteristics of the creatures and guide the decision-making process step by step.}

    \label{fig:example_programs}
\end{figure}

\subsection{Limitations}
\label{sec:app.limitations}
We acknowledge the following limitations, which merit further exploration in future studies.
It is important to note that these limitations pertain to the specific, simplified instantiation of the algorithms used in this preliminary study, rather than to the LSP framework itself:
\begin{itemize}[noitemsep, topsep=0pt, parsep=0pt, partopsep=0pt, leftmargin=*]
    \item \textbf{Domain-Specific Language Design:} A common practice in NSp is to design DSLs suitable for specific tasks. This work presents only a basic example of a DSL designed for predictive tasks. Investigating a variety of DSL designs could enable LSPs to excel across a broader range of applications.
    \item \textbf{Program Complexity:} Our search algorithm prioritizes accuracy without considering the complexity of the resulting programs, potentially leading to redundancies. The complexity of the learned programs could be reduced either through post-processing (akin to code cleaning) or by integrating complexity regularization during the search process.
\end{itemize}

\subsection{Societal Impact}
The development and deployment of interpretable predictive models using Large Language Models (LLMs) have significant societal implications.
By enhancing the transparency and interpretability of AI systems, our approach addresses critical concerns related to trust, accountability, and fairness of the decision making process.
These improvements are particularly valuable in high-stakes domains such as healthcare, finance, and legal decision-making, where understanding the rationale behind AI decisions is crucial for gaining user trust and ensuring ethical outcomes.

However, as with any AI technology, careful consideration must be given to the potential risks of misuse or unintended consequences. It is essential to continue developing comprehensive guidelines and regulatory frameworks to ensure that the deployment of these models aligns with societal values and ethical standards. By promoting transparency and interpretability, our approach paves the way for more responsible and beneficial integration of AI into society.

\subsection{License}
The open-source code from GitHub used in this paper adheres to various licenses like MIT, Apache 2.0, and GPL, ensuring the code's free use, modification, and distribution under specific conditions.
The ChatGPT API from OpenAI and the Gemini API from Google are used in compliance with their respective terms of service, which include usage restrictions, attribution requirements, and provisions for commercial use.
By following these licenses and terms, we maintain ethical and legal standards in utilizing both open-source code and proprietary APIs in our research.


\end{document}